\documentclass{article}

\usepackage{PRIMEarxiv}

\usepackage{hyperref}
\usepackage{url}

\usepackage{wrapfig}
\usepackage{graphicx}
\usepackage{subfigure}
\usepackage{nicefrac}
\usepackage{multirow}
\usepackage{booktabs,arydshln}

\usepackage{amssymb}

\usepackage{algorithm}
\usepackage[algo2e,ruled,linesnumbered]{algorithm2e}

\usepackage{xcolor}

\makeatletter
\g@addto@macro\normalsize{%
	\setlength\abovedisplayskip{.2ex}
	\setlength\belowdisplayskip{.15ex}
	\setlength\abovedisplayshortskip{.2ex}
	\setlength\belowdisplayshortskip{.15ex}
}
\makeatother

\makeatletter
\def\adl@drawiv#1#2#3{%
        \hskip.5\tabcolsep
        \xleaders#3{#2.5\@tempdimb #1{1}#2.5\@tempdimb}%
                #2\z@ plus1fil minus1fil\relax
        \hskip.5\tabcolsep}
\newcommand{\cdashlinelr}[1]{%
  \noalign{\vskip\aboverulesep
           \global\let\@dashdrawstore\adl@draw
           \global\let\adl@draw\adl@drawiv}
  \cdashline{#1}
  \noalign{\global\let\adl@draw\@dashdrawstore
           \vskip\belowrulesep}}
\makeatother

\title{Does Continual Learning Equally Forget All Parameters?
}

\author{
Haiyan Zhao$^1$\And
Tianyi Zhou$^2$\And
Guodong Long$^1$\And
Jing Jiang$^1$\And
Chengqi Zhang$^1$\and
$^1$University of Technology Sydney\\
$^2$University of Maryland\\
Haiyan.Zhao-2@student.uts.edu.au,
zhou@umiacs.umd.edu,\\
\{guodong.long, jing.jiang, Chengqi.Zhang\}@uts.edu.au
}

\begin{document}
\maketitle

\begin{abstract}
Distribution shift (e.g., task or domain shift) in continual learning (CL) usually results in catastrophic forgetting of neural networks. Although it can be alleviated by repeatedly replaying buffered data, the every-step replay is time-consuming.
In this paper, we study which modules in neural networks are more prone to forgetting by investigating their training dynamics during CL. Our proposed metrics show that only a few modules are more task-specific and sensitively alter between tasks, while others can be shared across tasks as common knowledge. Hence, we attribute forgetting mainly to the former and find that finetuning them only on a small buffer at the end of any CL method can bring non-trivial improvement.
Due to the small number of finetuned parameters, such ``Forgetting Prioritized Finetuning (FPF)'' is efficient 
in computation.
We further propose a more efficient and simpler method that entirely removes the every-step replay and replaces them by only $k$-times of FPF periodically triggered during CL. 
Surprisingly, this ``$k$-FPF'' performs comparably to FPF and outperforms the SOTA CL methods but significantly reduces their computational overhead and cost. 
In experiments on several benchmarks of class- and domain-incremental CL, FPF consistently improves existing CL methods by a large margin, and $k$-FPF further excels in efficiency without degrading the accuracy. 
We also empirically studied the impact of buffer size, epochs per task, and finetuning modules on the cost and accuracy of our methods.
\end{abstract}


\section{Introduction}
Empowered by advancing deep learning techniques and neural networks, machine learning has achieved unprecedented promising performance on challenging tasks in different fields, mostly under the i.i.d. offline setting. However, its reliability and performance degenerate drastically in continual learning (CL) where the data distribution or task in training changes over time, as the model quickly adapts to a new task and overwrites the previously learned weights. This leads to a severe bias toward more recent tasks and ``catastrophic forgetting'' of previously learned knowledge, which is detrimental to a variety of practical applications.

A widely studied strategy to mitigate forgetting is experience replay (ER)~\cite{ratcliff1990connectionist,robins1995catastrophic} and its variants~\cite{riemer2018learning,buzzega2020dark,boschini2022class}, which store a few data from previous tasks in the limited memory and train the model using both the current and buffered data. However, they only bring marginal improvements when the memory is too small to store sufficient data for recovering previously learned knowledge, which is common due to the complicated distributions of previous tasks. 
In contrast, multi-task learning~\cite{caruana1997multitask} usually adopts a model architecture composed of a task-agnostic backbone network and multiple task-specific adaptors on top of it. While the backbone needs to be pre-trained on large-scale data, the adaptors are usually lightweight and can be achieved using a few data. 
In CL, however, we cannot explicitly pre-define and separate the task-agnostic parts and task-specific parts.
Although previous methods~\cite{schwarz2018progress,zenke2017continual} have studied to restrict the change of parameters critical to previous tasks, a such extra constraint might degrade the training performance and discourage task-agnostic modules capturing shared knowledge.  

In this paper, we study a fundamental but open problem in CL, i.e., are most parameters task-specific and sensitively changing with the distribution shift? Or is the catastrophic forgetting mainly caused by the change in a few task-specific parameters? It naturally relates to the plasticity-stability trade-off in biological neural systems~\cite{mermillod2013stability}: more task-specific parameters improve the plasticity but may cause severe forgetting, while the stability can be improved by increasing parameters shared across tasks. In addition, how many task-specific parameters suffice to achieve promising performance on new task(s)? Is every-step replay necessary?  

To answer these questions, we investigate the training dynamics of model parameters during the course of CL by measuring their changes over time. For different CL methods training on different neural networks, we consistently observe that \textbf{only a few parameters change more drastically than others between tasks}. The results indicate that most parameters can be shared across tasks, and \textbf{we only need to finetune a few task-specific parameters to retain the previous tasks' performance}. Since these parameters only contain a few layers of various network architectures, they can be efficiently and accurately finetuned using a small buffer.

The empirical studies immediately motivate a simple yet effective method, ``forgetting prioritized finetuning (FPF)'' which finetunes the task-specific parameters using buffered data at the end of CL methods. 
Surprisingly, on multiple datasets, FPF consistently improves several widely-studied CL methods and substantially outperforms a variety of baselines.
Moreover, we extend FPF to a \textbf{more efficient replay-free CL method} ``$k$-FPF'' that entirely eliminates the cost of every-step replay by replacing such frequent replay with occasional FPF. $k$-FPF applies FPF only $k$ times during CL. We show that a relatively small $k$ suffices to enable $k$-FPF to achieve comparable performance with that of FPF+SOTA CL methods and meanwhile significantly reduces the computational cost.
In addition, we explore different groups of parameters to finetune in FPF and $k$-FPF by ranking their sensitivity to task shift evaluated in the empirical studies. 
For FPF, we compare them under different choices for the buffer size, the number of epochs per task, the CL method, and the network architecture. FPF can significantly improve existing CL methods by only finetuning $\leq \textbf{1.13\%}$ parameters. 
For $k$-FPF, we explore different groups of parameters, $k$, and the finetuning steps per FPF. $k$-FPF can achieve a promising trade-off between efficiency and performance.
Our experiments are conducted on a broad range of benchmarks for class- and domain-incremental CL in practice, e.g., medical image classification and realistic domain shift between image styles.\looseness-1

\section{Related Work}

\textbf{Continual Learning and Catastrophic Forgetting}
A line of methods stores samples of past tasks to combat the forgetting of previous knowledge.
ER~\cite{riemer2018learning} applies reservoir sampling~\cite{vitter1985random} to maintain a memory buffer of uniform samples over all tasks.
MIR~\cite{aljundi2019online} proposes a new strategy to select memory samples suffering the largest loss increase induced by the incoming mini-batch, so those at the forgetting boundary are selected.
DER and DER++~\cite{buzzega2020dark} apply knowledge distillation to mitigate forgetting by storing the output logits for buffered data during CL.
iCaRL~\cite{rebuffi2017icarl} selects samples closest to the representation mean of each class and trains a nearest-mean-of-exemplars classifier to preserve the class information of samples.
Our methods are complementary techniques to these memory-based methods. It can further improve their performance by finetuning a small portion of task-specific parameters on buffered data once (FPF) or occasionally ($k$-FPF).

Another line of work imposes a regularization on model parameters or isolates task-specific parameters to retain previous knowledge.
oEWC~\cite{schwarz2018progress} constrains the update of model parameters important to past tasks by a quadratic penalty.
To select task-specific parameters, SI~\cite{zenke2017continual} calculates the effect of the parameter change on the loss while MAS~\cite{aljundi2018memory} calculates the effect of the parameter change on the model outputs when each new task comes. 
PackNet~\cite{mallya2018packnet} and HAT~\cite{serra2018overcoming} iteratively assign a subset of parameters to consecutive tasks via binary masks.
All these works try to identify critical parameters for different tasks during CL and restrict the update of these parameters. But they can also prevent task-agnostic parameters from learning shared knowledge across tasks. From the training dynamics of CL, we identify the parameters sensitive to distribution shift. FPF and $k$-FPF finetune these parameters to mitigate bias without restricting the update of task-agnostic parameters.

\textbf{Different modules in neural networks}

Pham et al. \cite{pham2022continual} and Lesort et al. \cite{lesort2021continual} only study the effect of different normalization layers and classifiers on CL in a given setting, while our method investigates the sensitivity of all parameters in different network architectures and scenarios.
Wu et al. \cite{wu2022pretrained} study the forgetting of different blocks in the pre-trained language models by investigating their representation ability. We provide a more fine-grained analysis of the forgetting of each module by their training dynamics. And we find that parameters of different kinds of modules have different sensitivities to forgetting.
Ramasesh et al. \cite{ramasesh2020anatomy} show that freezing earlier layers after training the first task has little impact on the performance of the second task. This is because their unfrozen part covers the last FC layer and many BN parameters, which are the most sensitive/critical according to our empirical study.
Zhang et al. \cite{zhang2019all}
find that in different architectures, the parameters in the top layers(close to input) are more critical and perturbing them leads to poor performance. Our empirical study is consistent with their findings in that the earlier convolutional layer is sensitive to task drift and the induced biases on them lead to catastrophic forgetting.

\section{Problem Setup}

\textbf{Notations}
We consider the CL setting, where the model is trained on a sequence of tasks indexed by $t \in \{1,2,\dots,T\}$. During each task $t$, the training samples $(x,y)$ (with label $y$) are drawn from an i.i.d. distribution $D_t$. 
Given a neural network $f_{\Theta}(\cdot)$ of $L$ layers with parameter $\Theta=\{\theta_{\ell}\}_{\ell=1:L}$,  $\theta_\ell=\{\theta_{\ell,i}\}_{i=1:n_{\ell}}$ denote all parameters in layer-$\ell$ where $\theta_{\ell,i}$ denotes parameter-$i$. On each task, $f_{\Theta}(\cdot)$ is trained for $N$ epochs. We denote all parameters and the layer-$\ell$'s parameters at the end of the $n$-th epoch of task $t$ by $\Theta^{t}_{n}$ and $\theta^{t}_{\ell,n}$, $n\in\{1,\dots,N\}$, respectively.

\textbf{Settings}
In this paper, we mainly focus on class-incremental learning (class-IL) and domain-incremental learning (domain-IL). In class-IL, $D_t$ are drawn from a subset of classes $C_t$, and $\{C_t\}_{t=1}^T$ for different tasks are assumed to be disjoint.
Class-IL is a more challenging setting of CL\cite{van2019three} than task-incremental learning (task-IL)~\cite{lopez2017gradient}.
Unlike task-IL, class-IL cannot access the task label during inference and has to distinguish among all classes from all tasks. 
In domain-IL, tasks to be learned remain the same, but the domain varies, i.e. the input data distribution $D_t$ changes. The model is expected to adapt to the new domain without forgetting the old ones.
The goal of the class-IL and domain-IL is:
$\min_\Theta L(\Theta) \triangleq \sum_{t=1}^{T}\mathbb{E}_{(x,y)\sim D_t}[l(y,f_{\Theta}(x))]$, where $l$ is the objective function.\looseness-1

\textbf{Datasets}
We conduct class-IL experiments on Seq-MNIST, Seq-OrganAMNIST, Seq-PathMNIST, Seq-CIFAR-10, and Seq-TinyImageNet.
Seq-OrganAMNIST and Seq-PathMnist are generated by splitting OrganAMNIST or PathMNIST from MedMNIST\cite{medmnistv2}, a medical image classification benchmark. CL on medical images is essential in practice but also challenging since medical images always come as a stream with new patients and new diseases. Moreover, medical images of different classes might only have subtle differences that are hard to distinguish.
For domain-IL experiments, we use PACS dataset~\cite{li2017deeper}, which is widely used for domain generalization. It can present a more realistic domain-shift challenge than the toy-setting of PermuteMNIST~\cite{kirkpatrick2017overcoming}. Details of these datasets can be found in Appendix.~\ref{sec:dataset_detail}.

\textbf{Models}
We follow the standard network architectures adopted in most previous CL works.
For Seq-MNIST, following~\cite{
riemer2018learning}, we employ an MLP, i.e., a fully-connected (FC) network with two hidden layers, each composed of 100 ReLU units.
Following~\cite{
li2020sequential,derakhshani2022lifelonger}, we train ResNet-18~\cite{he2016deep} on other five datasets.
In addition, we also extend our empirical study to another architecture, i.e., VGG-11~\cite{simonyan2014very} on Seq-CIFAR-10.\looseness-1

\section{Forgetting of Different Parameters: An Empirical study}

A fundamental and long-lasting question in CL is how the distribution shift impacts different model parameters and why it leads to harmful forgetting. 
Its answer could unveil the plasticity-stability trade-off in CL, where some parameters are plastic and task-specific and thus have to be finetuned before deploying the model, while the stable ones can be shared with and generalized to new tasks. 
In order to answer this question, we conduct a comprehensive empirical study that compares the training dynamics of different parameters in three widely studied neural networks.
\looseness-1


\begin{figure*}[htp]
     \centering
     \subfigure[VGG(consecutive epochs)]{
         \centering
         \includegraphics[width=0.3\textwidth]{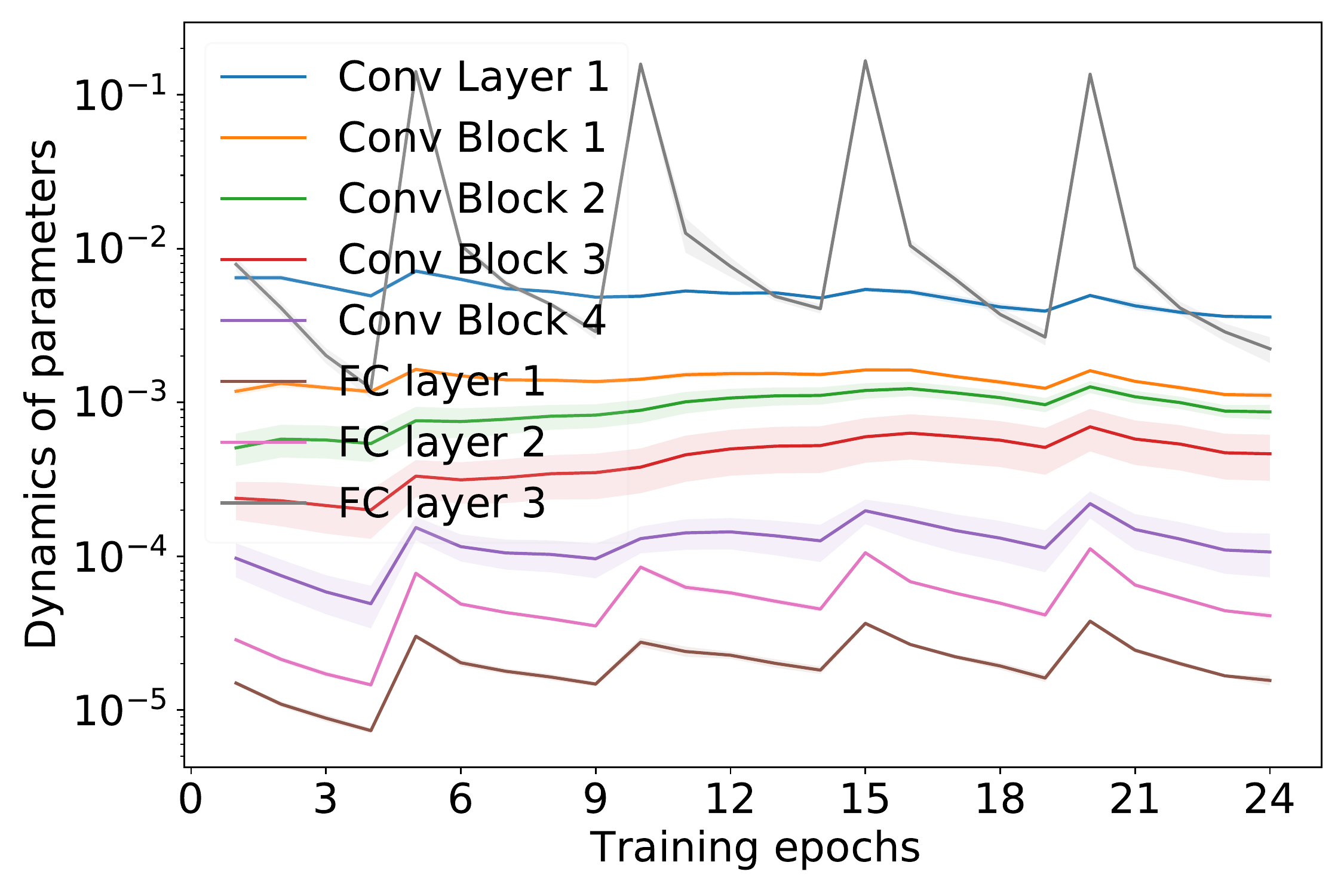}
     }
     \subfigure[VGG(consecutive tasks)]{
         \centering
         \includegraphics[width=0.3\textwidth]{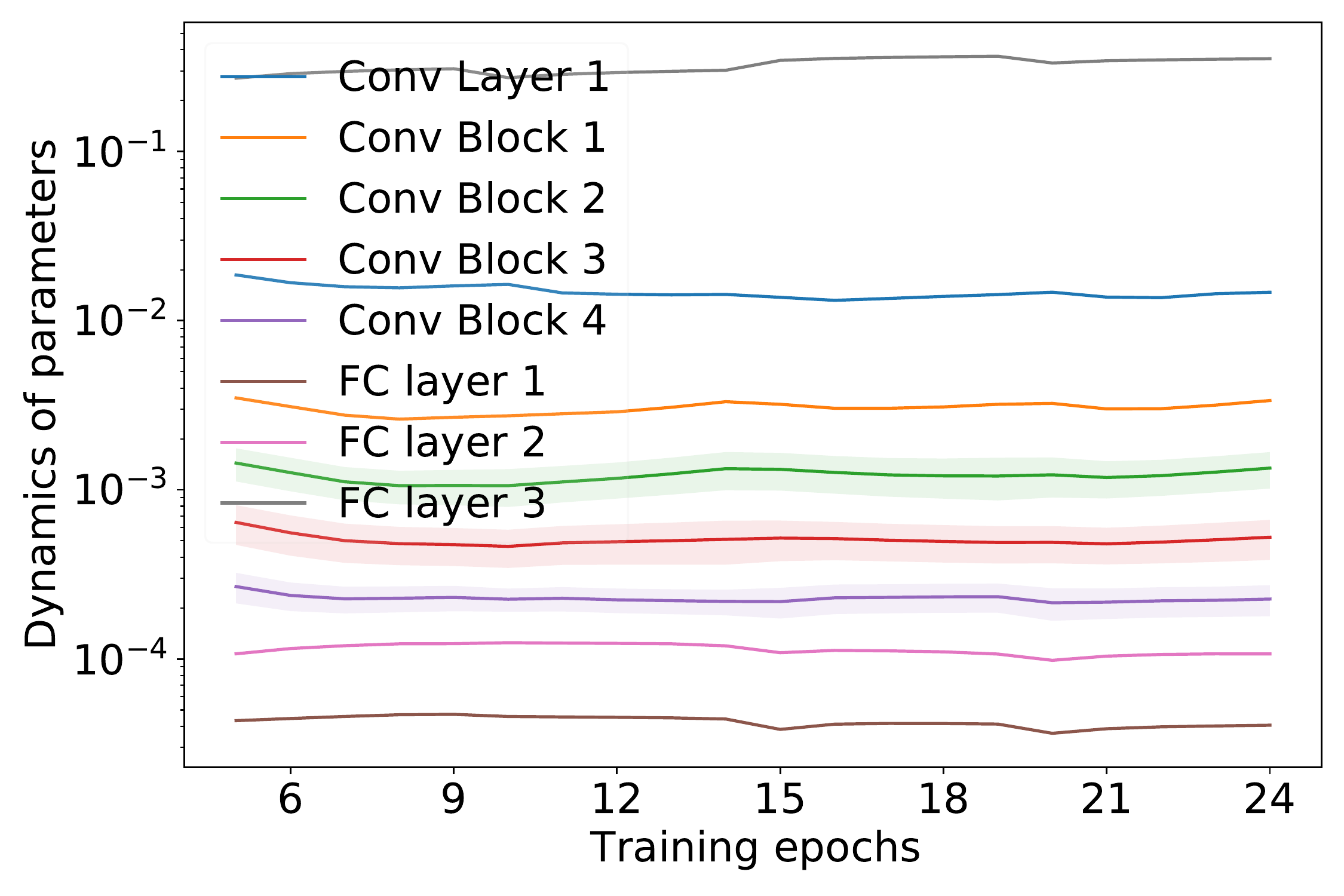}
     }
     \subfigure[ResNet(consecutive epochs)]{
         \centering
         \includegraphics[width=0.3\textwidth]{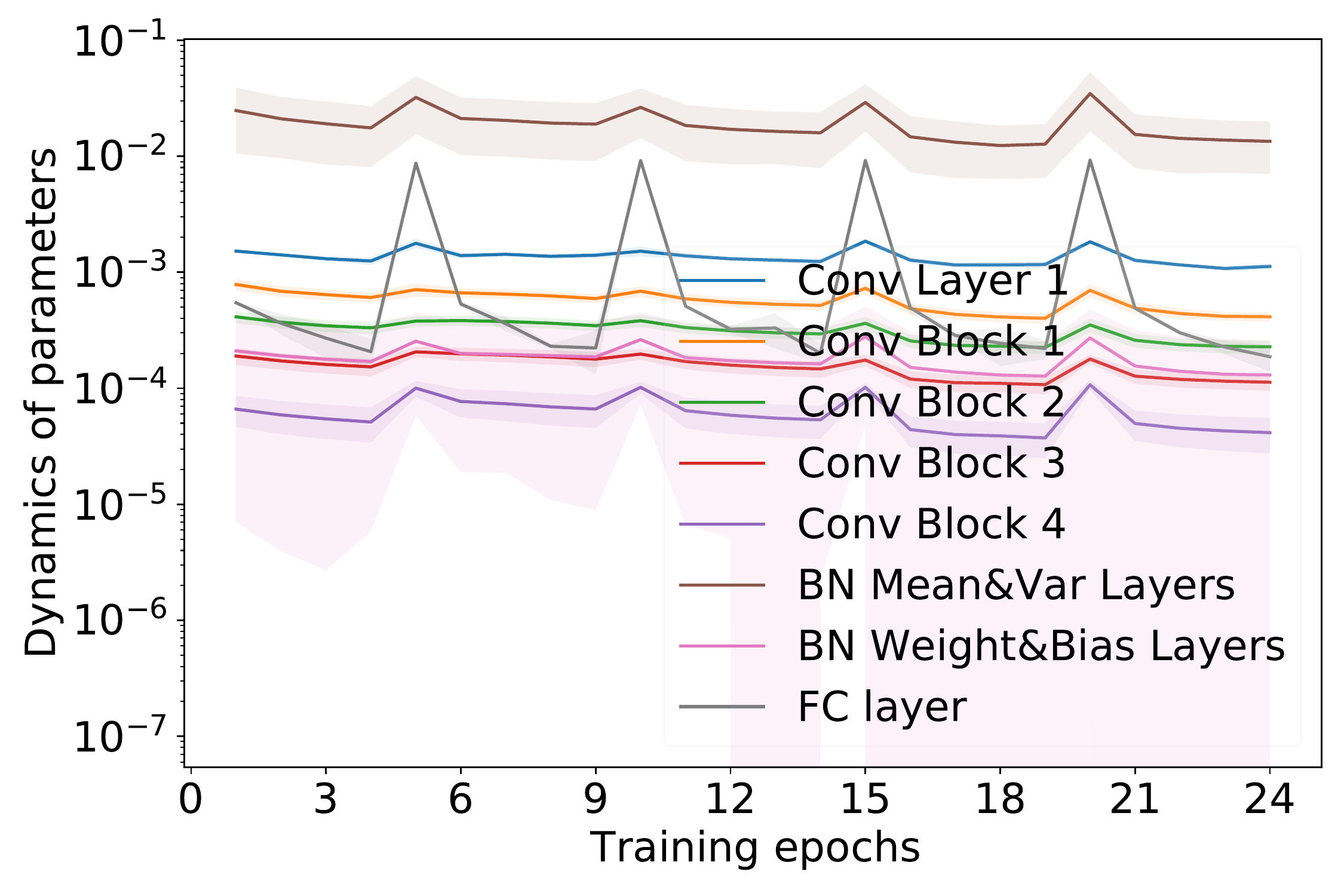}
     }
     \subfigure[ResNet(consecutive tasks)]{
         \centering
         \includegraphics[width=0.3\textwidth]{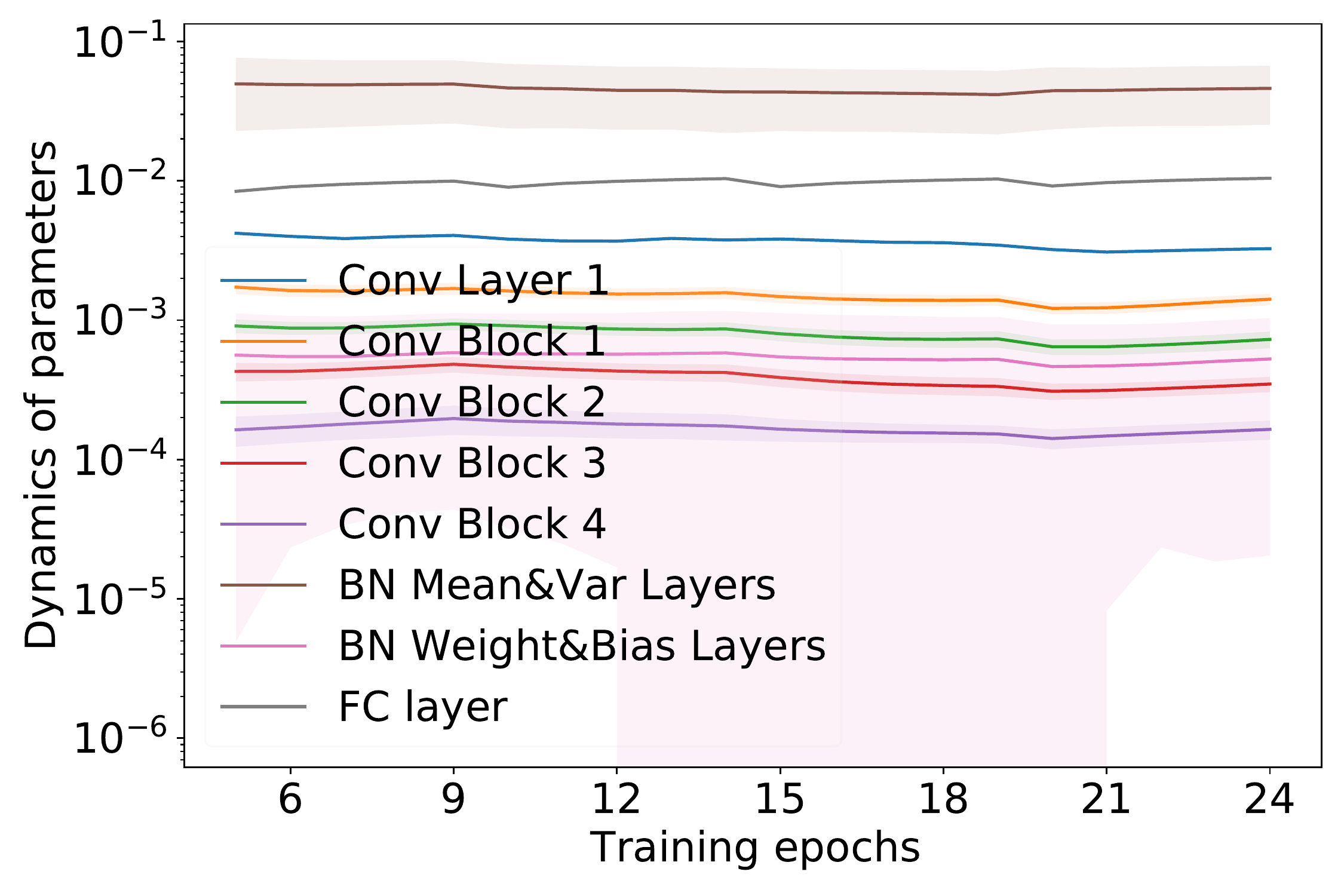}
     }
     \subfigure[MLP]{
         \centering
         \includegraphics[width=0.3\textwidth]{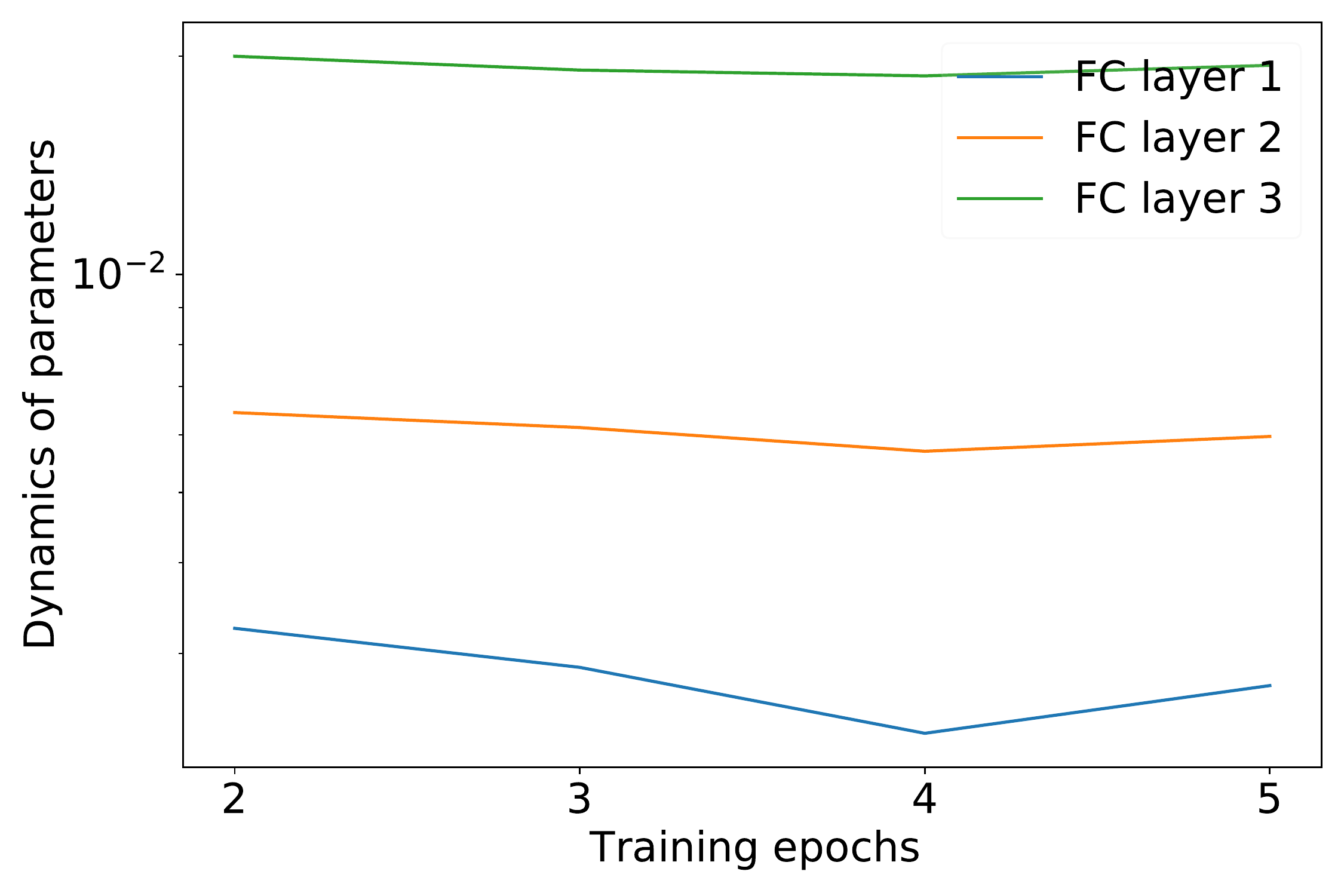}
     }
     \subfigure[Seq-OrganAMNIST]{
         \centering
         \includegraphics[width=0.3\textwidth]{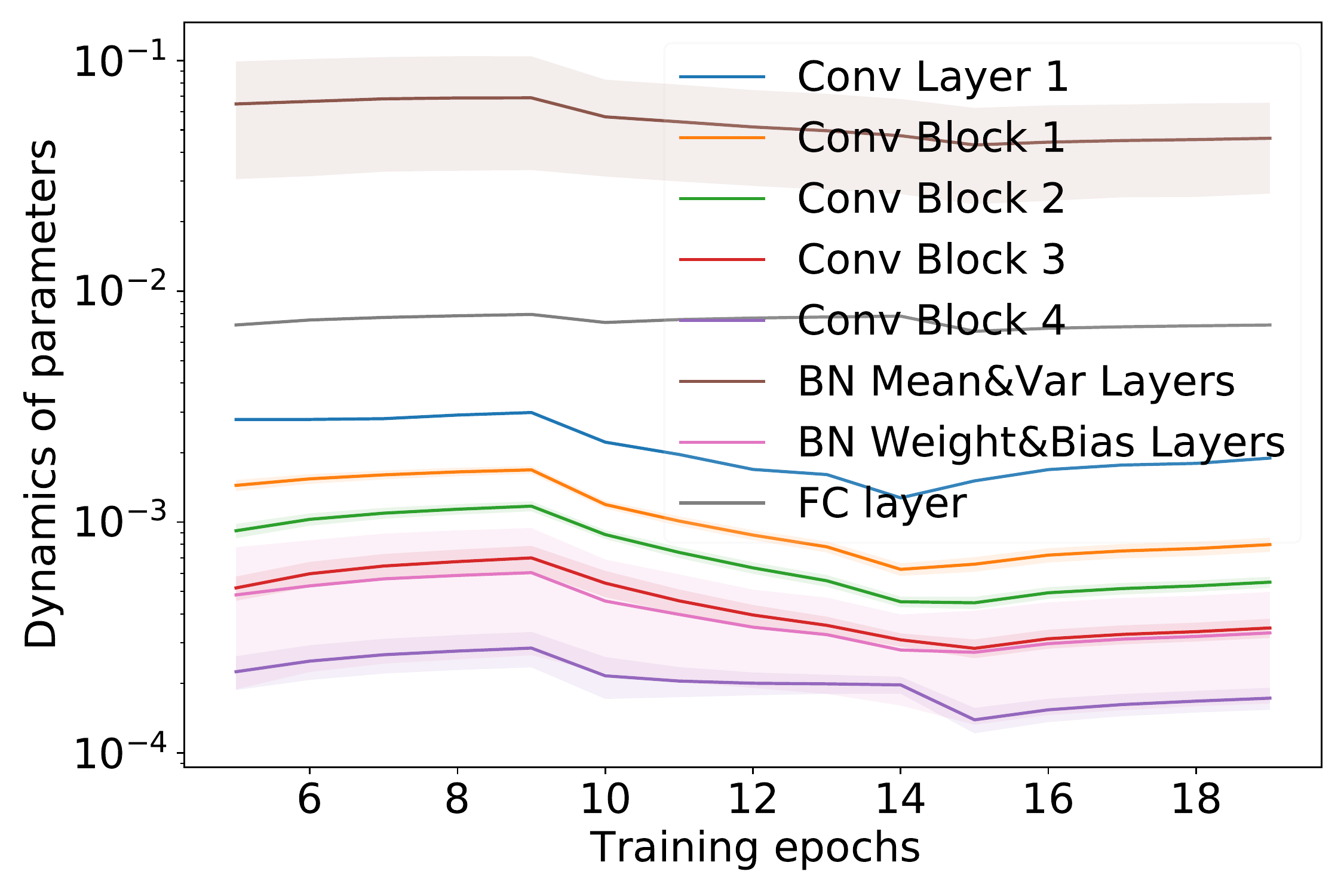}
     }
     \subfigure[ER(buffer size$=2000$)]{
         \centering
         \includegraphics[width=0.3\textwidth]{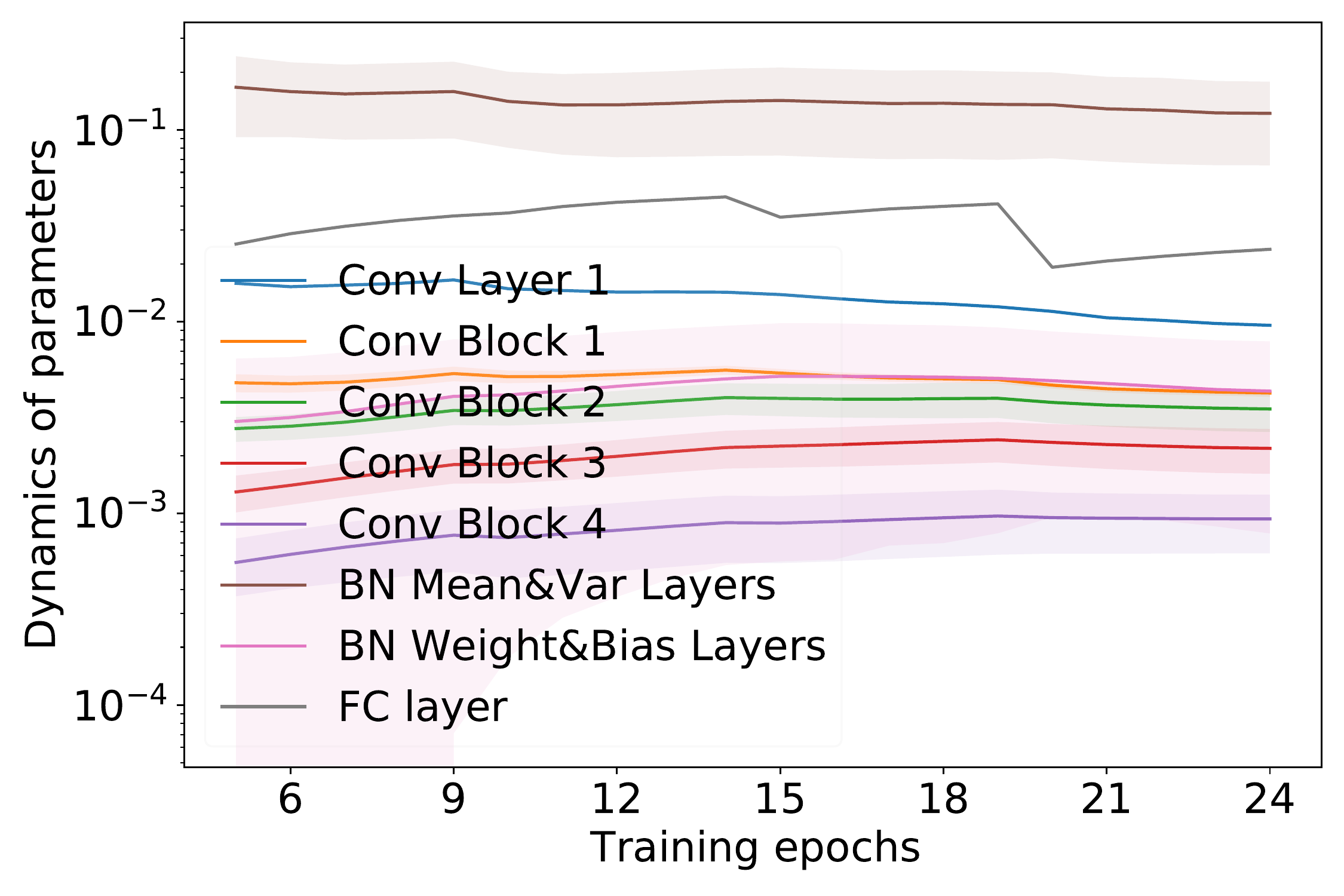}
     }
     \subfigure[Seq-PACS]{
         \centering
         \includegraphics[width=0.3\textwidth]{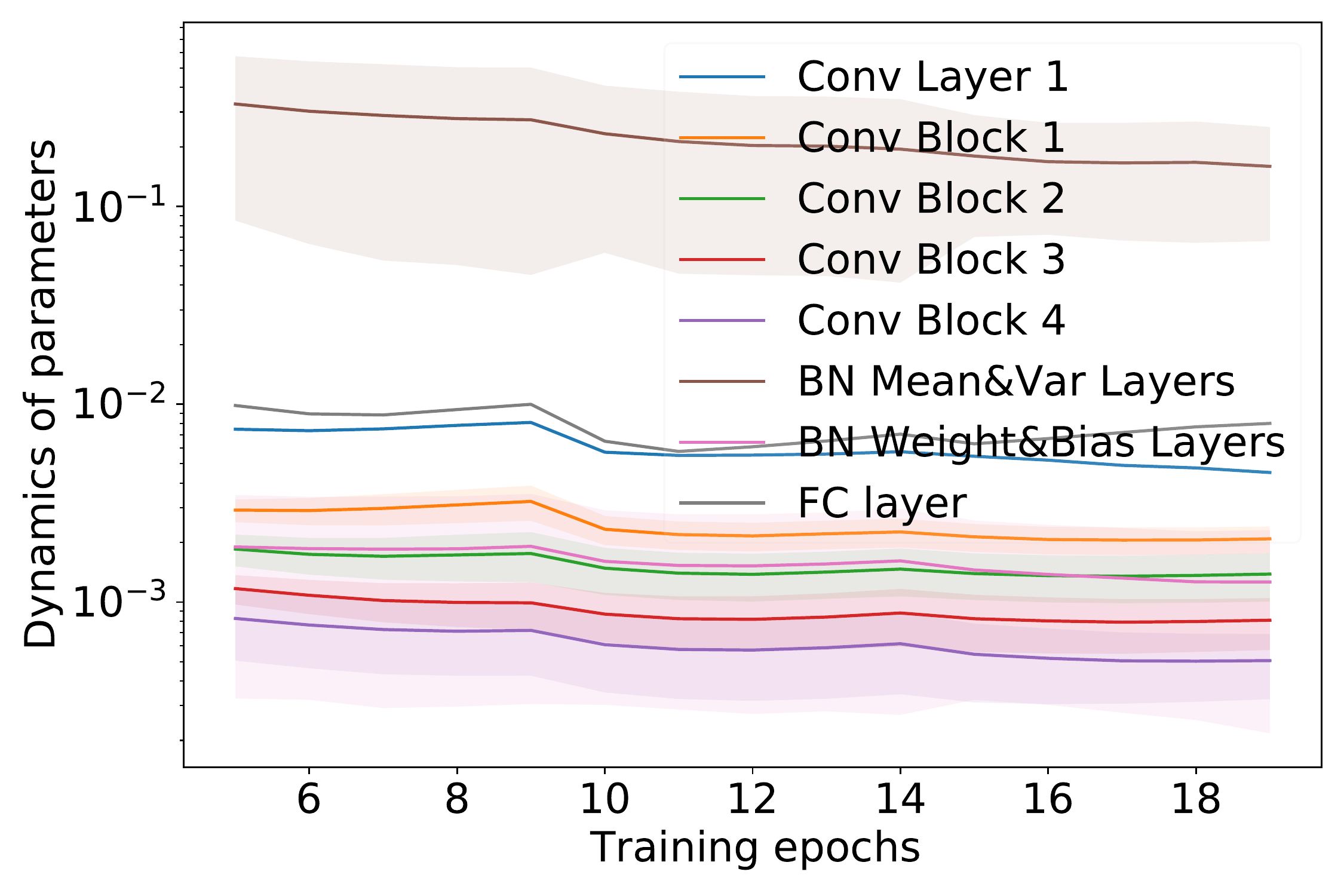}
     }
\caption{\footnotesize
(a-e) The training dynamics of different metrics for different groups of parameters when applying SGD in CL to train three types of deep neural networks; The training dynamics of other scenarios in ResNet-18:
(e) on a non-standard dataset; (g) using a different CL method with a different buffer size; (h) in the domain-IL setting. Note the y-axis is of logarithmic scale.}
\label{fig:more_dynamics}
\end{figure*}

\subsection{Measuring Forgetting via Training Dynamics}\label{sec:metric}

To measure and compare the forgetting effects of different parameters, we adopt two intuitive metrics to compute the change of parameters and investigate their dynamics over CL. 
The first metric calculates the difference between two consecutive epochs, e.g., for parameter $\theta_\ell$, it computes (1) $(1/|\theta_\ell|)\|\theta^t_{\ell,n} - \theta^t_{\ell,n-1}\|_1$ 
between epoch-$(n-1)$ and epoch-$n$ within a task-$t$ and (2) $(1/|\theta_\ell|)\|\theta^{t+1}_{\ell,1} - \theta^t_{\ell,N}\|_1$
between the last epoch of task-$t$ and the first epoch of task-$(t+1)$.
The training dynamics of this metric on different groups of parameters for different networks are shown in plots (a,c) of Fig.~\ref{fig:more_dynamics}.
In CL, the unstable changes of parameters are mainly caused by the task shift, while the learning within each task usually leads to smooth changes. Hence, the second metric focuses on the difference between two consecutive tasks, e.g., the change of parameters between epoch-$n$ of the two tasks, i.e., 
$C_\ell=(1/|\theta_\ell|)\|\theta^{t+1}_{\ell,n} - \theta^t_{\ell,n}\|_1$. Its results on different neural networks are displayed in plots (b,d,e) of Fig.~\ref{fig:more_dynamics}.

\subsection{Forgetting of Different Parameters During CL}

We first investigate and compare the training dynamics of different parameters in three types of networks. 
To gain insights applicable to all CL methods, we exclude any specific CL techniques but simply apply SGD to train a model on a sequence of tasks without any countermeasure to forgetting. Then, we extend the experiment to different CL methods and datasets to verify whether the observations still hold.\looseness-1

\textbf{Dynamics between Consecutive Epochs}
Plots (a,c) of Fig.~\ref{fig:more_dynamics} show the training dynamics of consecutive epochs for parameters in VGG-11 and ResNet-18 when trained on Seq-CIFAR10.
We partition all parameters in VGG-11 into several groups, i.e., the bottom convolutional layer (closest to the input), convolutional layers in different blocks, and three FC layers. 
Besides the groups of VGG-11, ResNet-18 applies batch-normalization (BN), which has two groups of parameters, i.e., (1) weights and biases and (2) mean and variance.
In the plots, all parameters experience more changes at the epoch of task switching and quickly converge after a few epochs in the same task. Hence, the dynamic patterns of this metric can be used to detect task boundaries. 

\textbf{Dynamics between Consecutive Tasks}
Plots (b,d,e) of Fig.~\ref{fig:more_dynamics} show the training dynamics of consecutive tasks for parameters in VGG-11, ResNet-18 and MLP.
We train a three-layer MLP for Seq-MNIST. Since each task in Seq-MNIST is trained only $1$ epoch, the dynamics of MLP for consecutive epochs and consecutive tasks are the same.
From the plots of different neural networks, the last FC layer is more sensitive to task shift than other layers. When BN is included in the network, BN layers' mean and variance become the most changed parameters. These observations are similar to studies in research domains like multi-task learning and domain adaptation~\cite{long2015learning,chang2019domain} that the last FC layer and BN layers are task-specific and cannot be shared among tasks. In CL, the last FC layer is sensitive because tasks in class-IL differ on their predicted classes, which are the outputs of the last FC layer. It is also intuitive that BN layers are task-specific since the mean and variance of BN layers capture the first and second-order moments of the distribution for the latent representations. The variance of BN weight and bias is relatively large compared to other layers. Please refer to Appendix.~\ref{sec:bn_dynamics} for details.\looseness-1

One interesting observation of VGG-11 and ResNet-18 is that the sensitivity of convolutional layers increases as the layer gets closer to the input. The reason may be that they are producing representations for the input images, whose distribution shift directly impacts the bottom convolutional layer. The functionality of filters in the top layers is to integrate patterns learned in the bottom layers to produce high-level patterns, so filters in the top layers are relatively stable. In Fig.~\ref{fig:filter_dynamics} of Appendix, we further study the training dynamics of each filter within a task or cross tasks in different layers of a network. Firstly, the training dynamics of each filter in the bottom layer are much larger than that of the top layer, which is in line with our above observation. We also find that in the same layer, when tasks shift, the dynamics of a small number of filters increase significantly. These filters should be task-specific and critical to the learning of the new task.\looseness-1

\textbf{Dynamics on different scenarios}
The above empirical study is limited to SGD without applying any other CL techniques and only focuses on class-IL. In the following, we extend the studies to different CL methods, non-standard datasets, and domain-IL while fixing the model to be ResNet-18.
Fig.~\ref{fig:more_dynamics} (f) extends the empirical study to a medical dataset Seq-OrganAMNIST. Compared to Seq-CIFAR-10, it differs in the number of tasks, dataset size, image size, and data type.
We further replace SGD with ER using replay buffer, whose results are reported in Fig.~\ref{fig:more_dynamics} (g). 
The ranking order of parameter groups in terms of sensitivity stays consistent under the change of dataset and replay strategy.\looseness-1

In domain-IL, as shown in Fig.~\ref{fig:more_dynamics} (h), the training dynamics of different parameters are in line with our observations in class-IL: only a small portion of parameters are task-specific.
However, one difference is worth noting. Since the output classes stay the same across tasks and only the input domain changes, the last FC layer which is the most sensitive in class-IL, becomes equally or less sensitive than the bottom convolutional layer.
Hence, the plasticity and stability of parameters are impacted by how close they are to the changed data distributions.

\textbf{Inspiration from Empirical Studies.} Above studies shed light on the improvements of CL methods. \textbf{(1)} We compare the sensitivity of different parameters in three types of deep neural networks and observe that only a small portion of them are much more sensitive than others. This implies that only finetuning these task-specific (or plastic) parameters may suffice to retain the previous tasks. 
\textbf{(2)} The dynamics between consecutive epochs show that all layers experience more changes when tasks shift, which can be used to detect task boundaries during CL. Knowing task boundaries is a critical prerequisite for lots of CL methods. The proposed metric makes the CL problem much easier, makes these methods more general and can contribute to better CL methods. \textbf{(3)} According to Fig.~\ref{fig:filter_dynamics}, in convolutional layers, only a mall part of task-specific filters leads to the great change of dynamics when tasks shift. The regularization or isolation of these filters can improve the performance of CL.\looseness-1

\begin{figure*}[ht]
     \centering
     \subfigure{
         \centering
         \includegraphics[width=0.8\textwidth]{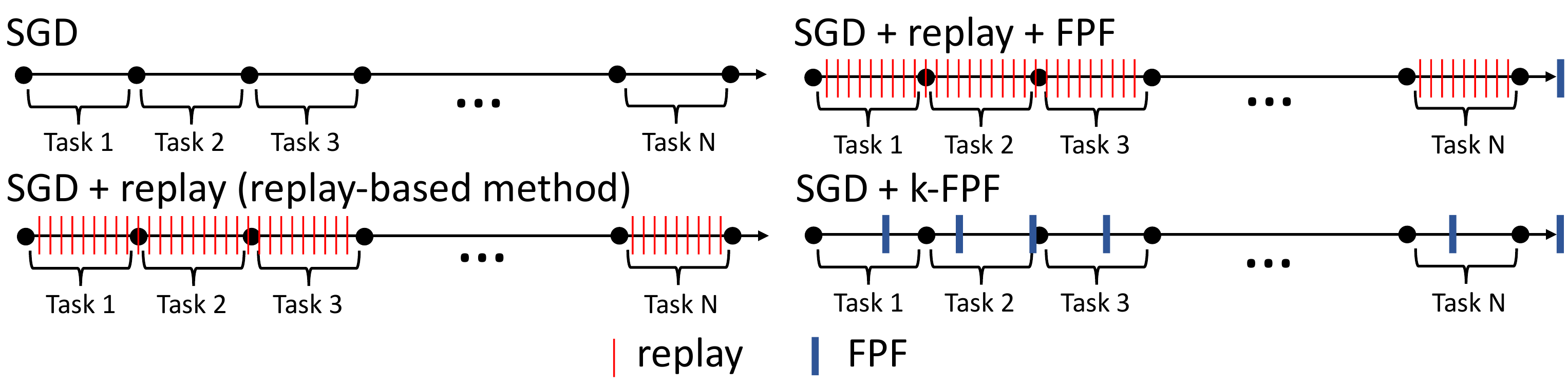}
     }
\caption{\footnotesize Comparison of SGD, replay-based method, FPF and $k$-FPF.
SGD trains tasks sequentially without replay.
Replay-based methods train models on buffered and current data simultaneously. 
FPF finetunes the most sensitive parameters for a few iterations using buffered data at the end of arbitrary CL methods. $k$-FPF periodically (regardless of task boundaries) applies FPF for $k$ times during training.}
\label{fig:FPF_flows}
\end{figure*}

\section{Forgetting Prioritized Finetuning Methods}
The above empirical study of the training dynamics on parameters immediately motivates a simple but novel method for CL, i.e., ``forgetting prioritized finetuning (FPF)'', which can be applied to any existing CL method. In the more efficient $k$-FPF, we further remove the every-step replay and any other CL techniques but simply applies $k$-times of FPF in SGD training. In Fig.~\ref{fig:FPF_flows}, we provide an illustration that compares SGD, replay-based methods and our methods. At last, we propose a metric to automatically identify sensitive parameters in each neural network. See Appendix.~\ref{sec:detailed_alg} for the detailed procedure of FPF and $k$-FPF.\looseness-1

\textbf{FPF to improve CL performance.}
FPF applies lightweight finetuning to the most task-specific parameters using the buffered data after the training of arbitrary CL methods. Hence, it is complementary to any existing CL methods as a correction step to remove their biases in the task-specific parameters by finetuning them on unbiased buffered data. Thereby, it can improve the performance of any existing CL methods without causing notably extra computation.

\textbf{$k$-FPF to improve CL efficiency and performance.}
FPF is a simple technique that brings non-trivial improvement, but it is applied after the training of an existing CL method. Unfortunately, many SOTA CL methods require time-consuming replay in every step, which at least doubles the total computation.
Since only a few parameters are sensitive during the task shift, can we develop a replay-free and lazy CL that replaces every-step replay with occasional FPF? 
We propose $k$-FPF that applies FPF $k$ times during CL as shown in Fig.~\ref{fig:FPF_flows}. Without the costly experience replay, $k$-FPF can still achieve comparable performance as FPF+SOTA CL methods but only requires nearly half of their computation.
We can apply $k$-FPF with any replay-free method, e.g., SGD, which
is usually used as a lower-bound for CL methods. We still maintain a small buffer by reservoir sampling, but it is only for FPF, so SGD never accesses it. We lazily apply FPF on the buffer after every $\tau$ SGD step (in total $k$ times over $k\tau$ SGD steps) without knowing the task boundaries.\looseness-1

\textbf{$k$-FPF-CE+SGD} We propose two variants of $k$-FPF, i.e., $k$-FPF-CE+SGD and $k$-FPF-KD+SGD. 
$k$-FPF-CE+SGD uses the cross-entropy loss to update the sensitive parameters during each FPF. In this paper, $k$-FPF-CE refers to $k$-FPF-CE+SGD if not specified. The objective of FPF in $k$-FPF-CE is: $\min_{\Theta^{\star}} L(\Theta^{\star}) \triangleq \mathbb{E}_{(x,y)\sim B}[l_{CE}(y,f_{\Theta}(x))]$
where $\Theta^{\star}$ denotes selected groups of task-specific parameters, $B$ refers to the buffered data and $l_{CE}$ is the cross-entropy loss.

\textbf{$k$-FPF-KD+SGD to further improve performance} Inspired by DER~\cite{buzzega2020dark}, we further propose $k$-FPF-KD that introduces knowledge distillation (KD)~\cite{hinton2015distilling} to the objective in $k$-FPF-CE. In this paper, $k$-FPF-KD refers to $k$-FPF-KD+SGD if not specified. Same as DER, the pre-softmax responses (i.e. logits) for buffered data at training time are stored in the buffer as well. During FPF, the current model is trained to match the buffered logits to retain the knowledge of previous models. The objective of FPF in $k$-FPF-KD is: 
$\min_{\Theta^{\star}} L(\Theta^{\star}) \triangleq \mathbb{E}_{(x,y)\sim B}[l_{CE}(y,f_{\Theta}(x))] + \lambda \mathbb{E}_{(x,z)\sim B}[l_{MSE}(z,h_{\Theta}(x))]$
where $z$ is the logits of buffered sample $x$, $l_{MSE}$ refers to the mean-squared loss, $h_\Theta(x)$ computes the pre-softmax logits and $\lambda$ is a hyper-parameter balancing the two terms.
Compared to the computation of $k$-FPF-CE, the additional computation of $k$-FPF-KD is negligible.

\begin{table*}[ht]
\caption{\footnotesize
Test accuracy ($\%$) of baselines, FPF and $k$-FPF.
``-'' indicates that the algorithm is not applicable to the setting.
$k$-FPF-KD applies an additional knowledge distillation loss to the objective of $k$-FPF-CE.
\textbf{Bold} and \textcolor{gray}{\underline{\textbf{Bold gray}}} mark the best and second best accuracy.\looseness-1}
\begin{center}
\begin{small}
\begin{sc}
\resizebox{0.9\textwidth}{!}{
\begin{tabular}{cllllll}
\toprule
\multirow{2}*{\textbf{ Buffer}} & \multirow{2}*{\textbf{Methods}} & \multicolumn{4}{c}{\textbf{class-IL}} & \multicolumn{1}{c}{\textbf{Domain-IL}} \\
&  & Seq-OrganAMNIST& Seq-PathMNIST & Seq-CIFAR-10 & Seq-Tiny-ImageNet&  Seq-PACS \\
\midrule
& JOINT & 91.92$\pm$0.46 & 82.47$\pm$2.99 & 81.05$\pm$1.67 & 41.57$\pm$0.55 & 70.85$\pm$8.90\\
& SGD   & 24.19$\pm$0.15 & 23.65$\pm$0.07 & 19.34$\pm$0.06 & 7.10$\pm$0.14 & 31.43$\pm$6.39   \\
& oEWC~\cite{schwarz2018progress}  & 22.71$\pm$0.67 & 22.36$\pm$1.18 & 18.48$\pm$0.71 & 6.58$\pm$0.12 & 35.96$\pm$4.59  \\
\midrule
\multicolumn{1}{c}{\multirow{14}*{200}}
& GDUMB~\cite{prabhu2020gdumb}                 & 61.78$\pm$2.21 & 46.31$\pm$5.64 & 30.36$\pm$2.65 & 2.43$\pm$0.31  & 34.16$\pm$3.45 \\
& \textbf{$k$-FPF-CE}           & 75.21$\pm$2.03 & \textcolor{gray}{\underline{\textbf{72.88$\pm$3.22}}} & 57.97$\pm$1.53 & 13.76$\pm$0.72 & 60.70 $\pm$2.81\\
& \textbf{$k$-FPF-KD}   & \textcolor{gray}{\underline{\textbf{80.32$\pm$1.16}}} & \textbf{74.68$\pm$4.72} & 58.50$\pm$1.03 & \textcolor{gray}{\underline{\textbf{14.74$\pm$0.94}}} & 63.15$\pm$1.19\\
\cdashlinelr{2-7}
& ER~\cite{riemer2018learning}                    & 71.69$\pm$1.71 & 51.66$\pm$5.86 & 45.71$\pm$1.44 & 8.15$\pm$0.25  & 51.53$\pm$5.10 \\ 
& \textbf{FPF}+ER            & 76.92$\pm$2.26 & 67.34$\pm$2.68 & 57.68$\pm$0.76 & 13.08$\pm$0.65 & \textcolor{gray}{\underline{\textbf{65.16$\pm$1.97}}} \\
\cdashlinelr{2-7}
& AGEM~\cite{chaudhry2018efficient}                  & 24.16$\pm$0.17 & 27.93$\pm$4.24 & 19.29$\pm$0.04 & 7.22$\pm$0.15  & 40.54$\pm$3.43 \\
& \textbf{FPF}+AGEM          & 72.22$\pm$2.45 & 66.88$\pm$3.05 & 55.33$\pm$2.19 & 12.27$\pm$0.49 & 57.33$\pm$0.76 \\
\cdashlinelr{2-7}
& iCaRL~\cite{rebuffi2017icarl}                 & 79.61$\pm$0.56 & 54.35$\pm$0.94 & 59.60$\pm$1.06 & 12.13$\pm$0.20 & - \\
& \textbf{FPF}+iCaRL         & 80.28$\pm$0.58 & 71.20$\pm$2.19 & \textbf{63.36$\pm$0.91} & \textbf{16.99$\pm$0.37} & - \\
\cdashlinelr{2-7}
& FDR~\cite{benjamin2018measuring}                   & 68.29$\pm$3.27 & 44.27$\pm$3.20 & 41.77$\pm$4.24 & 8.81$\pm$0.19  & 45.91$\pm$3.54 \\
& \textbf{FPF}+FDR           & 76.10$\pm$0.87 & 70.06$\pm$2.78 & 51.91$\pm$2.77 & 11.52$\pm$0.72 & 57.17$\pm$1.31 \\
\cdashlinelr{2-7}
& DER~\cite{buzzega2020dark}                   & 73.28$\pm$1.33 & 54.45$\pm$5.92 & 47.04$\pm$3.03 & 9.89$\pm$0.58  & 46.93$\pm$4.94  \\
& \textbf{FPF}+DER           & 79.63$\pm$1.21 & 67.29$\pm$3.75 & 56.67$\pm$2.19 & 12.65$\pm$0.60 & 61.49$\pm$1.37 \\
\cdashlinelr{2-7}
& DER++~\cite{buzzega2020dark}                 & 78.22$\pm$2.05 & 62.00$\pm$3.79 & 59.13$\pm$0.81 & 12.12$\pm$0.69 & 55.75$\pm$2.02  \\
& \textbf{FPF}+DER++         & \textbf{80.99$\pm$0.91} & 68.78$\pm$2.99 & \textcolor{gray}{\underline{\textbf{61.69$\pm$0.97}}} & 13.72$\pm$0.40 & \textbf{65.28$\pm$1.02} \\

\midrule
\multicolumn{1}{c}{\multirow{14}*{500}}
&  GDUMB~\cite{prabhu2020gdumb}                & 73.29$\pm$1.82 & 63.55$\pm$5.62 & 42.18$\pm$2.05 & 3.67$\pm$0.25  & 43.29$\pm$2.53 \\
& \textbf{$k$-FPF-CE}           & 81.28$\pm$0.71 & 76.72$\pm$1.94 & 64.35$\pm$0.87 & 19.57$\pm$0.37 & 65.90$\pm$0.72 \\
& \textbf{$k$-FPF-KD}   & 85.16$\pm$0.67 & \textbf{79.20$\pm$3.89} & 66.43$\pm$0.50 & \textbf{20.56$\pm$0.32} & \textcolor{gray}{\underline{\textbf{66.42$\pm$2.21}}}\\
\cdashlinelr{2-7}
& ER~\cite{riemer2018learning}                    & 80.45$\pm$0.99 & 57.54$\pm$3.05 & 57.64$\pm$4.27 & 10.09$\pm$0.34 & 52.72$\pm$4.01 \\
& \textbf{FPF}+ER            & 84.07$\pm$1.26 & 69.83$\pm$2.87 & 65.47$\pm$2.64 & 18.61$\pm$0.70 & 64.27$\pm$1.91 \\
\cdashlinelr{2-7}
& AGEM~\cite{chaudhry2018efficient}                  & 24.00$\pm$0.18 & 27.33$\pm$3.93 & 19.47$\pm$0.03 & 7.14$\pm$0.10  & 35.29$\pm$4.94  \\
& \textbf{FPF}+AGEM          & 78.98$\pm$1.80 & 73.32$\pm$3.73 & 57.84$\pm$1.98 & 16.16$\pm$0.30 & 62.40$\pm$1.89 \\
\cdashlinelr{2-7}
& iCaRL~\cite{rebuffi2017icarl}                 & 82.95$\pm$0.47 & 57.67$\pm$1.13 & 62.26$\pm$1.09 & 14.81$\pm$0.37 & - \\
& \textbf{FPF}+iCaRL         & 83.88$\pm$0.69 & 73.56$\pm$3.00 & \textcolor{gray}{\underline{\textbf{67.75$\pm$0.67}}} & 16.69$\pm$0.29 & - \\
\cdashlinelr{2-7}
& FDR~\cite{benjamin2018measuring}                   & 76.62$\pm$1.81 & 40.08$\pm$4.13 & 43.52$\pm$1.74 & 11.33$\pm$0.33 & 48.50$\pm$4.67 \\
& \textbf{FPF}+FDR           & 82.32$\pm$0.91 & 73.64$\pm$3.85 & 63.09$\pm$0.81 & 17.10$\pm$0.35 & 65.39$\pm$1.83 \\
\cdashlinelr{2-7}
& DER~\cite{buzzega2020dark}                   & 82.52$\pm$0.52 & 66.71$\pm$3.40 & 55.98$\pm$3.35 & 11.54$\pm$0.70 & 47.63$\pm$3.85 \\
& \textbf{FPF}+DER           & \textcolor{gray}{\underline{\textbf{85.18$\pm$0.39}}} & 74.13$\pm$3.12 & 67.52$\pm$0.83 & 17.34$\pm$0.53 & 65.69$\pm$1.66 \\
\cdashlinelr{2-7}
& DER++~\cite{buzzega2020dark}                 & 84.25$\pm$0.47 & 71.09$\pm$2.60 & 67.06$\pm$0.31 & 17.14$\pm$0.66 & 57.77$\pm$2.54 \\
& \textbf{FPF}+DER++         & \textbf{85.40$\pm$0.26} & \textcolor{gray}{\underline{\textbf{77.37$\pm$1.32}}} & \textbf{69.08$\pm$0.92} & \textcolor{gray}{\underline{\textbf{20.17$\pm$0.35}}} & \textbf{66.89$\pm$1.32} \\

\bottomrule
\end{tabular} 
}
\label{tab:main_results}
\end{sc}
\end{small}
\end{center}
\end{table*}

\textbf{Selection of sensitive parameters for FPF and $k$-FPF}

A key challenge in FPF and $k$-FPF is to select the task-specific parameters for finetuning.
Examples of the training dynamics for different layers of various networks are shown in plots (b,d,e) of Fig.~\ref{fig:more_dynamics}, and their ranking does not change over epochs. So we propose to select sensitive parameters for different neural networks according to their training dynamics in the early epochs.
Specifically, for each neural network, its layers are partitioned into $G$ groups as shown in Fig.~\ref{fig:more_dynamics}, we calculate the sensitive score $S_g$ for each group of layers in the neural network by
\begin{equation}
    S_\textsl{g} = \frac{(1/|\textsl{g}|)\sum_{\ell\in \textsl{g}}C_\ell}{\sum_{g=1}^{G}(1/|g|)\sum_{\ell\in g}C_\ell}*G
\label{equ:sel_paras}
\end{equation}
where $C_\ell$ is the training dynamics mentioned in Sec.~\ref{sec:metric}. We calculate the ratio of sensitivity for group $\textsl{g}$ over all $G$ groups in the network. Since each network consists of a different number of parameter groups, we multiply $G$ to rescale the sensitivity score.\looseness-1


In the experiments later, under different scenarios and on various benchmarks, we evaluate the performance of FPF and $k$-FPF when selecting different subsets of task-specific parameters. 
In a nutshell, finetuning parameters of higher sensitivity achieve more improvement, which is in line with our findings in empirical studies. 
FPF outperforms all baselines when parameter groups whose $S_\textsl{g}$ $>1$ are regarded as sensitive parameters and account for only $1.13\%, 0.32\%$ and $0.15\%$ of the number of all parameters in MLP, VGG-11 and ResNet-18.
For $k$-FPF, finetuning more parameters, i.e., the earlier convolutional layers, achieves the best performance. This is the price of removing replay, which halves the computational cost. We set the threshold of sensitivity score to $0.3$ so that $k$-FPF achieves SOTA performance and only $12.40\%, 1.69\%$ and $24.91\%$ of parameters in MLP, VGG-11 and ResNet-18 are regarded as sensitive parameters. \looseness-1


\section{Experiments}

In this section, to compare FPF and $k$-FPF with SOTA CL methods, we conduct our experiments mainly on ResNet-18.  
We apply FPF and $k$-FPF to multiple benchmark datasets and compare them with SOTA CL baselines in terms of test accuracy and efficiency. 
Besides, we also compare the performance of finetuning different parameters in FPF and $k$-FPF and show that finetuning a small portion of task-specific parameters suffices to improve CL.
FPF improves SOTA CL methods by a large margin under all these scenarios, while $k$-FPF achieves comparable performance with FPF but is more efficient.
Please refer to the Appendix for more results and analysis.

\textbf{Implementation Details.}
We follow the settings in \cite{buzzega2020dark} to train various SOTA CL methods on different datasets, except training each task for only $5$ epochs, which is more practical than $50$ or $100$ epochs in \cite{buzzega2020dark} for the streaming setting of CL. 
Since the epochs are reduced, we re-tune the learning rate and hyper-parameters for different scenarios by performing a grid-search on a validation set of $10\%$ samples drawn from the original training set.
For both FPF and $k$-FPF, we use the same optimizer, i.e., SGD with the cosine-annealing learning rate schedule, and finetune the selected parameters with a batchsize of 32 for all scenarios. The finetuning steps for FPF and $k$-FPF are $300$ and $100$, respectively.
We perform a grid-search on the validation set to tune the learning rate and other hyper-parameters.
Please refer to Appendix.~\ref{sec:parameter_search} for the hyper-parameters we explored. 

\textbf{Baseline methods.}
We apply FPF to several SOTA memory-based CL methods: ER, iCaRL, A-GEM~\cite{chaudhry2018efficient}, FDR~\cite{benjamin2018measuring}, DER and DER++. Besides, we also compare our methods with GDUMB~\cite{prabhu2020gdumb} and oEWC. 
We take JOINT as the upper bound for CL, which trains all tasks jointly, and SGD as the lower bound, which trains tasks sequentially without any countermeasure to forgetting.
For FPF, $k$-FPF, and all memory-based methods, the performance with buffer sizes 200 and 500 are reported.
All results reported in Tab.~\ref{tab:main_results} are averaged over five trials with different random seeds.

\begin{figure*}[htbp]
     \centering
      \subfigure[Seq-PathMNIST]{
         \centering
         \includegraphics[width=0.3\textwidth]{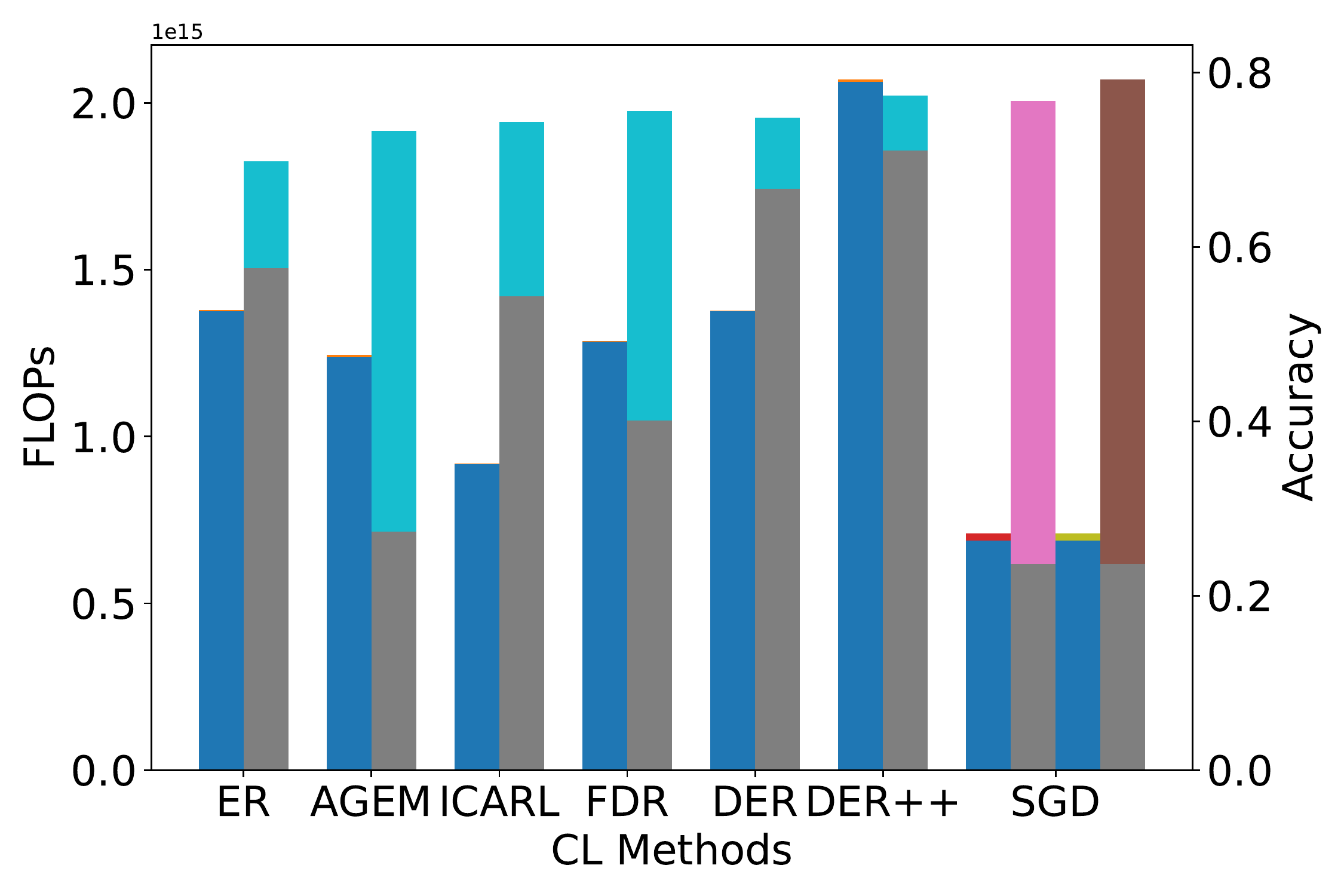}
     }
     \subfigure[Seq-Tiny-ImageNet]{
         \centering
         \includegraphics[width=0.3\textwidth]{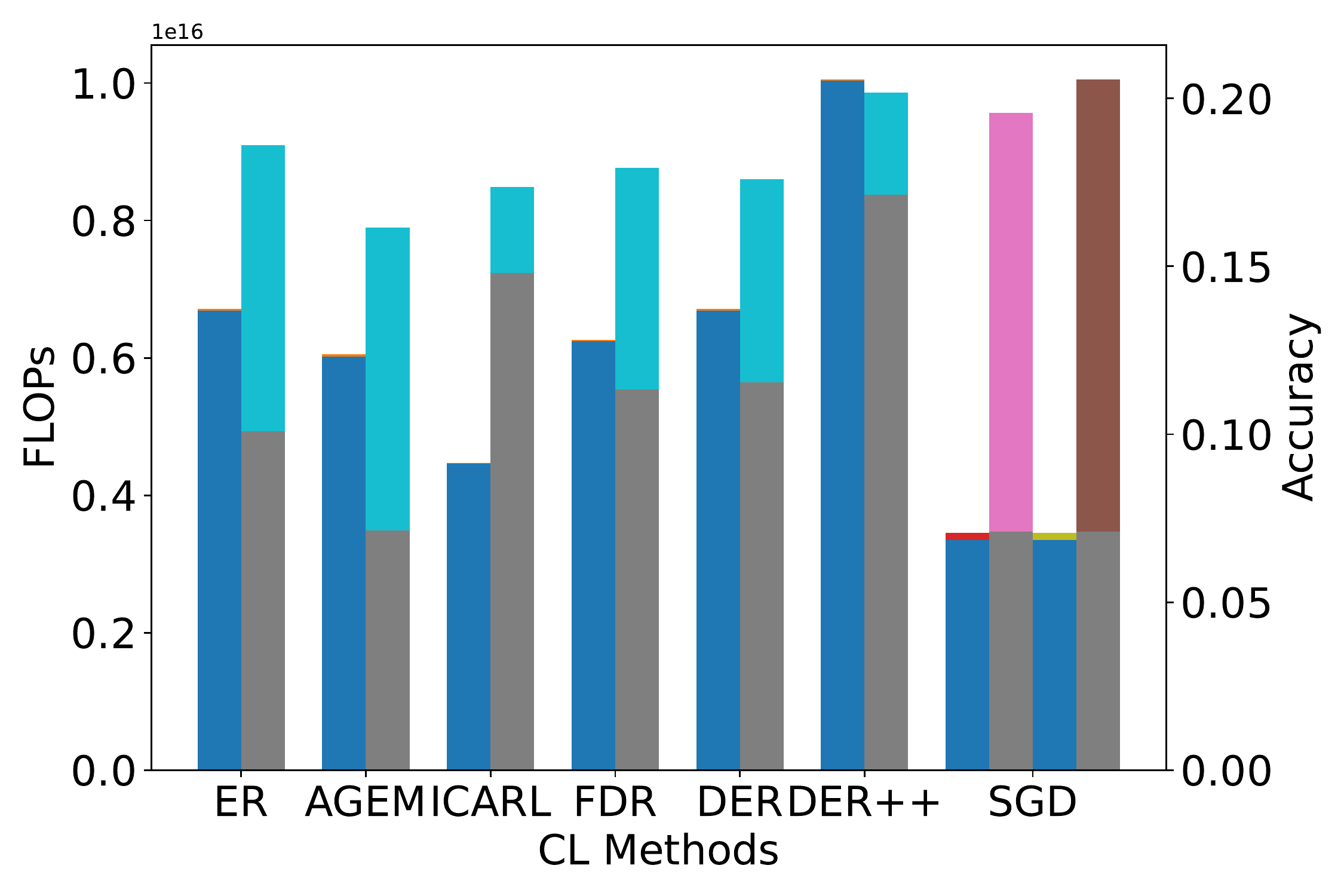}
     }
     \subfigure[Seq-PACS]{
         \centering
         \includegraphics[width=0.3\textwidth]{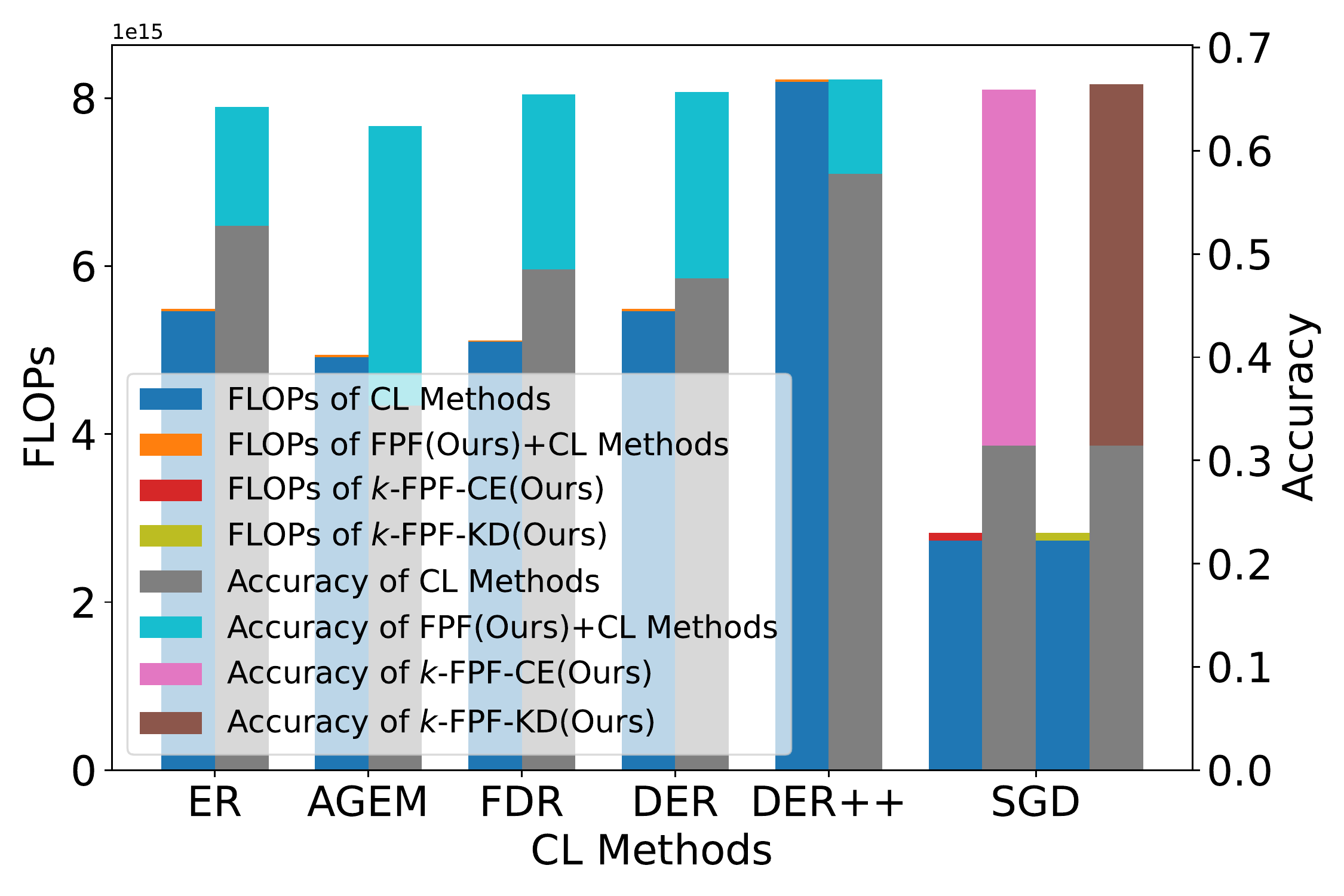}
     }
\caption{\footnotesize Comparison of FLOPs and accuracy between FPF, $k$-FPF and SOTA methods. \textbf{FPF improves all CL methods by a large margin without notably extra computation. $k$-FPF consumes much less computation but achieves comparable performance as FPF.}  A large and clear version can be found in Appendix.~\ref{sec:clear_img}.\looseness-1
}
\label{fig:methods_flops}
\end{figure*}

\subsection{Main Results}

\textbf{FPF considerably improves the performance of all memory-based CL methods} and achieves SOTA performance over all scenarios in class-IL and domain-IL in Tab.~\ref{tab:main_results}.
For methods with catastrophic forgetting, like AGEM, the accuracy of FPF increases exponentially.
The surge in performance illustrates that FPF can eliminate bias by finetuning task-specific parameters to adapt to all seen tasks.

\textbf{$k$-FPF-CE replaces the costly every-step replay with efficient occasional FPF.} In Tab.~\ref{tab:main_results}, the performance of $k$-FPF-CE on Seq-PathMNIST, Seq-Tiny-ImageNet and Seq-PACS are better than the best CL methods, and its performance on Seq-OrganAMNIST and Seq-Cifar10 are also better than most CL methods, which implies that finetuning the task-specific parameters on a small number of buffer during SGD can help retain the previous knowledge and mitigate forgetting, \textbf{every-step replay is not necessary}. 
In Fig.~\ref{fig:methods_flops}, the number of training FLOPs and the accuracy of different methods are reported. Compared to the training FLOPs of several CL methods, the computation cost of FPF and $k$-FPF-CE is almost negligible. The overall training FLOPs of $k$-FPF-CE are still much less than SOTA CL methods while its performance is better, which shows the efficiency of $k$-FPF-CE.

\textbf{$k$-FPF-KD further improves the performance of $k$-FPF-CE to be comparable to FPF.} 
$k$-FPF-CE proposes the efficiency of CL methods, but its performance is a bit worse than that of FPF. One of the most differences between $k$-FPF and FPF is the experience replay during the training of CL. Inspired by DER, we propose $k$-FPF-KD, which uses knowledge distillation to match the outputs of previous models on buffered data, hence retaining the knowledge of previous tasks. The results of $k$-FPF-KD in Tab.~\ref{tab:main_results} show that it is comparable to FPF in most scenarios. 
Fig.~\ref{fig:methods_flops} shows that the FLOPs of $k$-FPF-KD are similar to $k$-FPF-CE but much less than other CL methods and FPF, and in some cases, it outperforms FPF.
\textbf{$k$-FPF-KD shows SOTA performance in both efficiency and accuracy}.

\begin{figure}
         \centering
         \includegraphics[width=0.9\columnwidth]{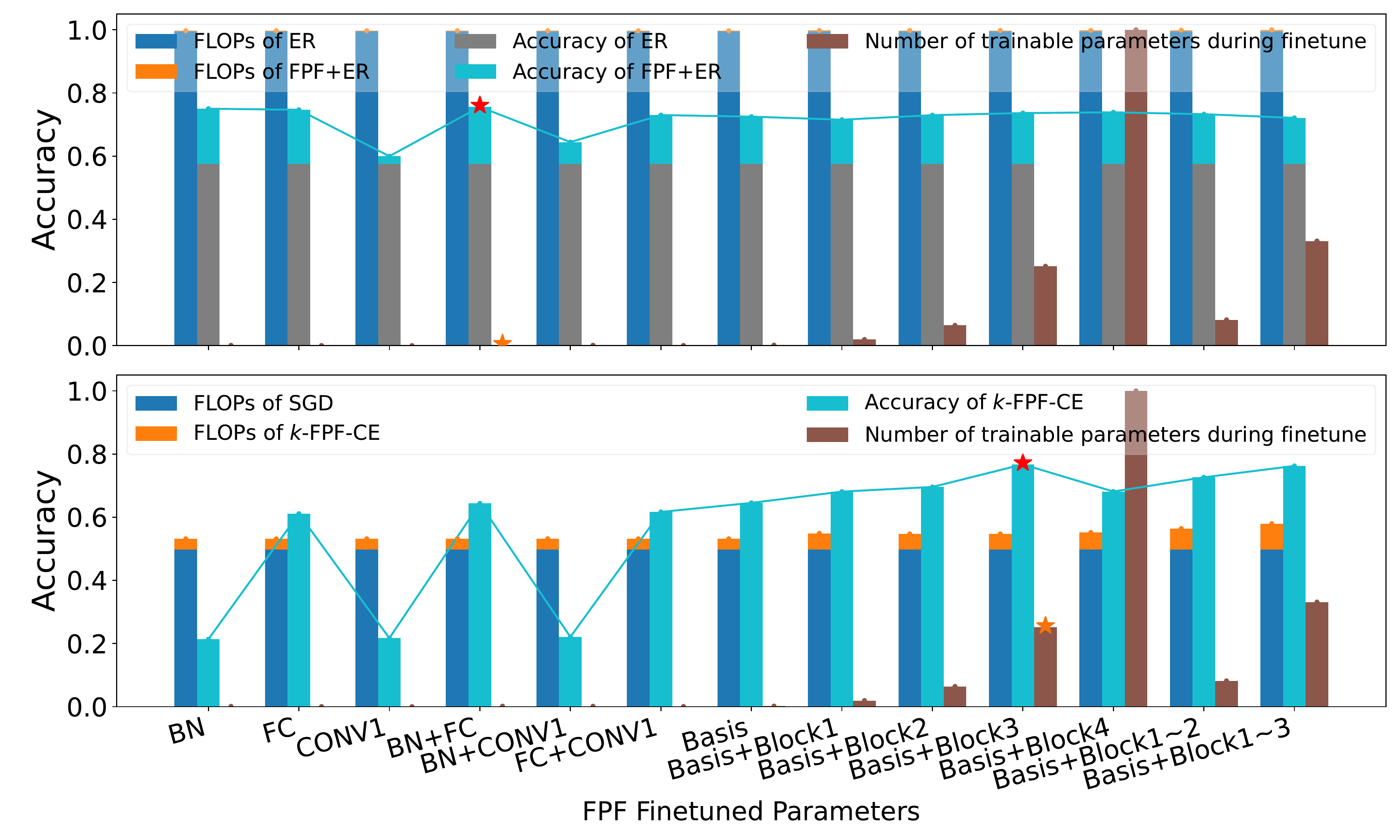}
\caption{\footnotesize Comparison of FLOPs, number of finetuned parameters, and accuracy for FPF(Top) and $k$-FPF(Bottom) finetuning different combinations of parameters. All FLOPs are normalized together to (0,1], as well as the number of finetuning parameters. ``Basis'' in x-label refers to ``BN+FC+CONV1''.
Red stars highlight the best accuracy and show \textbf{both FPF and $k$-FPF only require to finetune a small portion of task-specific parameters. $k$-FPF halves FPF's FLOPs.} A clear version can be found in Appendix.~\ref{sec:clear_img}.
}
\label{fig:ft_diffparts_flops}
\end{figure}

\subsection{Comparison of finetuning different parameters in FPF and $k$-FPF}\label{sec:efficiency}

\textbf{FPF and $k$-FPF get the best performance when only a small portion of task-specific parameters are finetuned.}
In Fig.~\ref{fig:ft_diffparts_flops}, the accuracy, training FLOPs and the number of trainable parameters during finetune of applying FPF or $k$-FPF to different task-specific parameters in ResNet-18 on Seq-PathMNIST are compared. Overall different scenarios, $k$-FPF only needs about half FLOPs of FPF with better performance (indicated by Red Stars).
When finetuning on different task-specific parameters, FPF performs the best when BN+FC layers are finetuned, which is only $0.127\%$ of all parameters (indicated by Orange Stars). This is consistent with our observations in empirical studies where BN and FC layers are the most sensitive parameters to distribution shift. And the results show that only finetuning a small portion of task-specific parameters can mitigate catastrophic forgetting and generalize the model.

The phenomenon for $k$-FPF is a little different. 
(1) In the bottom plot of Fig.~\ref{fig:ft_diffparts_flops}, when FC layer is not selected for finetuning in $k$-FPF, the performance is much worse. This is because, in class-IL, the output classes change across tasks, so the FC layer is trained to only output the classes for the current task~\cite{hou2019learning}. 
In contrast, when applying $k$-FPF to domain-IL on Seq-PACS, where the output classes keep the same for different tasks,  Fig.~\ref{fig:ft_diffparts_flops_pacs} in Appendix.~\ref{sec:domain_flops} shows that finetuning FC layer performs similarly as finetuning other parameters. Hence, the last FC layer is more sensitive in class-IL than in Domain-IL. This is also shown in Fig.~\ref{fig:more_dynamics} (d,h).
(2) As the red star indicates, $k$-FPF needs to finetune a little more parameters (Block3 of convolutional layers, $18.91\%$ of all parameters) to achieve a comparable accuracy with FPF. Without experience replay during SGD, the model has a larger bias on the current task, and thus more task-specific parameters are needed to be finetuned.
This also indicates that such bias of task-specific parameters is the main reason for catastrophic forgetting.
When Block4 ($75.22\%$ of all parameters) is finetuned, since it is the most stable group of parameters in our empirical study, the performance of $k$-FPF degrades.

\begin{figure*}[htbp]
     \centering
     \includegraphics[width=0.85\textwidth]{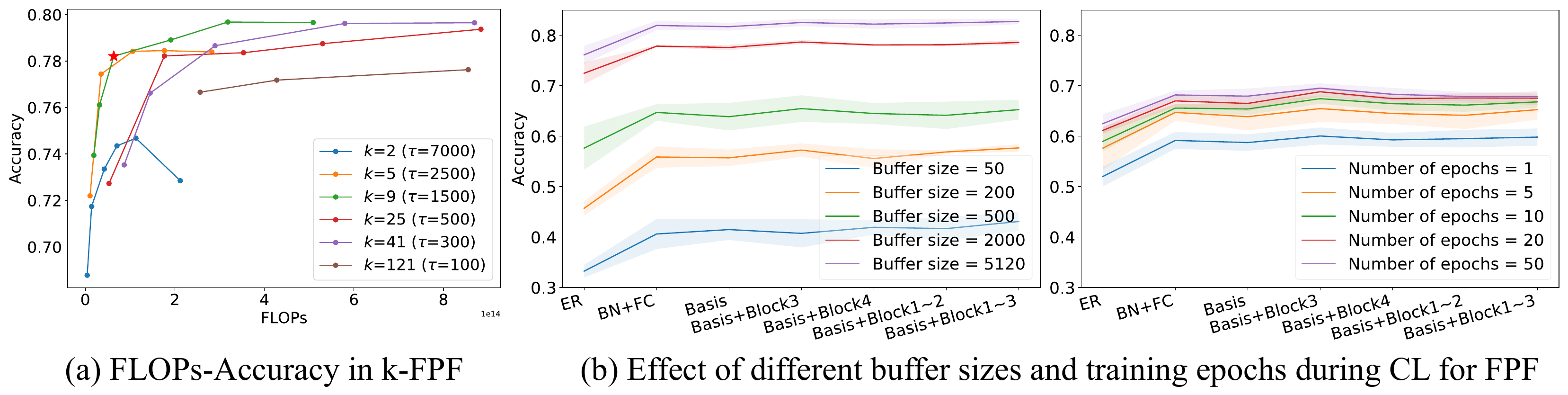}
\caption{\footnotesize \textbf{(a)} Trade-off between FLOPs and accuracy for $k$-FPF with different $k$ and $\tau$ (the SGD steps between two consecutive FPF). By increasing the finetuning steps per FPF, the accuracy quickly saturates. The best trade-off is highlighted at the top-left corner when $k=9 (\tau=1500)$. 
\textbf{(b)} Comparison between ER and FPF+ER finetuning different parameters with different buffer sizes and number of epochs per task. In all scenarios, FPF can significantly improve ER by only finetuning BN+FC.
}
\label{fig:FPFDT_different_intervals_epochs}
\end{figure*}

\subsection{Analysis of FPF and $k$-FPF in Different Scenarios}

\textbf{Different training FLOPs for $k$-FPF}
In Fig.~\ref{fig:FPFDT_different_intervals_epochs}(a), we study the trade-off between the training FLOPs and the accuracy of $k$-FPF on Seq-PathMNIST by changing $k$ and the number of finetuning steps. $\tau$ in the legend refers to the interval of two consecutive FPF.
Fixing $k$, $k$-FPF saturates quickly as the finetuning steps increase. This implies that $k$-FPF is efficient on FLOPs to achieve the best performance. 
For experiments with small $k$, e.g. $k$=2, though the computation required is very low, performance cannot be further improved. This implies that FPF needs to be applied on buffered samples more frequently to mitigate forgetting. When $k$ is large, e.g., $k$=41 or 121, the accuracy slightly improves with the price of much more required computation. As the red star in the plot indicates, applying FPF every $1500$ training steps can achieve the best computation-accuracy trade-off.

\textbf{Different buffer sizes and training epochs for FPF}
The buffer size and the training epochs per task are usually crucial in replay-based CL methods. In Fig.~\ref{fig:FPFDT_different_intervals_epochs}(b), when the buffer size or the number of epochs increases, the performance of ER improves as well. However, increasing the buffer size brings more benefits. When the buffer size or epochs grow too large, the performance of ER seems to saturate and increases slowly.
For all scenarios, finetuning BN+FC layers is highly effective in alleviating the current task's bias and promoting performance, which is consistent with our observations from the empirical studies.

\section{Conclusion}
We study a fundamental problem in CL, i.e., which parts of a neural network are task-specific and more prone to catastrophic forgetting. Extensive empirical studies in diverse settings  consistently show that only a small portion of parameters is task-specific and sensitive. This discovery leads to a simple yet effective ``forgetting prioritized finetuning (FPF)'' that only finetunes a subset of these parameters on the buffered data before model deployment. FPF is complementary to existing CL methods and can consistently improve their performance. We further replace the costly every-step replay with $k$-times of occasional FPF during CL to improve efficiency. Such $k$-FPF achieves comparable performance as FPF+SOTA CL while consuming nearly half of its computation. In future work, we will study how to further reduce the memory size required by FPF.\looseness-1

\bibliographystyle{unsrt}  
\bibliography{templateArxiv}

\newpage
\appendix
\onecolumn

\section{Detailed procedure of FPF and $k$-FPF}
\label{sec:detailed_alg}
The detailed algorithm of FPF, $k$-FPF and reservoir sampling are shown in Alg.~\ref{alg:FPF}, Alg.~\ref{alg:kFPF} and Alg.~\ref{alg:reservoir}. For FPF, if the existing CL method belongs to replay-based methods, the buffer will be reserved according to this method. Otherwise, FPF will reserve the buffer by reservoir sampling. For both FPF and $k$-FPF, the data of each task may be trained for more than one epoch.

\begin{algorithm}[H]
    \setcounter{AlgoLine}{0}
    \SetKwInOut{Input}{Input}
    \SetKwInOut{Output}{Output}
    \SetKwInOut{Init}{Initialize}
    \Input{ Dataset $D$, an existing CL method $M$, early epochs iteration $I$, number of finetuning iterations $K$}
    \Output{A well-trained CL model}
    \Init{Buffer $B\leftarrow \emptyset$, training iterations $i=0$}
    \For{($x$, $y$) $\in$ $D$}{
        Run a batch of CL method $M$;\par
        $i \leftarrow i+1 $;\par
        \If{$i$ = $I$}{
            Identify sensitive groups of parameters \{\textsl{g}\} by Eq.~\ref{equ:sel_paras}.
        }
    $B$ ← reservoir($B$, ($x$, $y$)).
    }
    Finetune the sensitive groups of parameters \{\textsl{g}\} for $K$ iterations on $B$.\par
\caption{\sc{FPF}}
\label{alg:FPF}
\end{algorithm}

\begin{algorithm}[H]
    \setcounter{AlgoLine}{0}
    \SetKwInOut{Input}{Input}
    \SetKwInOut{Output}{Output}
    \SetKwInOut{Init}{Initialize}
    \Input{ Dataset $D$, CL method SGD, early epochs iteration $I$, interval of FPF $\tau$, number of finetuning iterations $K$}
    \Output{A well-trained CL model}
    \Init{Buffer $B\leftarrow \emptyset$, training iterations $i=0$}
    \For{($x$, $y$) $\in$ $D$}{
        Run a batch of SGD on ($x$, $y$);\par
        $i \leftarrow i+1 $;\par
        \If{$i$ = $I$}{
            Identify sensitive groups of parameters \{\textsl{g}\} by Eq.~\ref{equ:sel_paras}.
        }
        \If{$i \% \tau = 0$}{
            Finetune the sensitive groups of parameters \{\textsl{g}\} for $K$ iterations on $B$.
        }
    $B$ ← reservoir($B$, ($x$, $y$)).
    }
    Finetune the sensitive groups of parameters \{\textsl{g}\} for $K$ iterations on $B$.\par
\caption{\sc{$k$-FPF}}
\label{alg:kFPF}
\end{algorithm}

\begin{algorithm}[H]
    \setcounter{AlgoLine}{0}
    \SetKwInOut{Input}{Input}
    \SetKwInOut{Output}{Output}
    \SetKwInOut{Init}{Initialize}
    \Input{Buffer $B$, number of seen examples $S$, example $x$, label $y$}
    \Output{An updated buffer $B$}

    \uIf{$|B| > S$}{
        $B[S] \leftarrow ($x$, $y$)$
    }
    \Else{
        $k =$ randomInteger(min = $0$, max = $S$));\par
        \If{$k < |B|$ }{
            $B[k] \leftarrow ($x$, $y$)$
        }
    }

\caption{\sc{Reservoir Sampling}}
\label{alg:reservoir}
\end{algorithm}


\clearpage
\section{Comparison between FPF and the method finetuning all parameters}

In Tab.\ref{tab:FPF_vs_FPFALL}, we compare FPF with FPF-ALL (which finetunes all parameters) when applied to different CL methods for two types of CL, i.e., class-IL and domain-IL. The results show that FPF consistently achieves comparable or slightly higher accuracy than FPF-ALL by spending significantly fewer FLOPs. This demonstrates the advantage of FPF on efficiency. 

\begin{table*}[h]
\caption{\footnotesize
Comparison of accuracy and FLOPs between FPF and FPF-ALL(finetuning all parameters).
}
\begin{center}
\begin{small}
\resizebox{0.75\textwidth}{!}{
\begin{tabular}{lcccc}
\toprule
\multirow{2}*{\textbf{ Methods}}  & \multicolumn{2}{c}{\textbf{Seq-PathMNIST}} & \multicolumn{2}{c}{\textbf{Seq-PACS}} \\
 & Accuracy & FLOPs(B) & Accuracy & FLOPs(B) \\
\midrule
\textbf{$k$-FPF-CE}    & 76.72$\pm$1.94 & 21.35 & 65.90$\pm$0.72 & 148.25\\
\textbf{$k$-FPF-ALL-CE}& 75.74$\pm$2.91 & 43.95 & 64.48$\pm$2.23 & 174.60 \\
\textbf{FPF}+ER         & 69.83$\pm$2.87 & 4.68 & 64.27$\pm$1.91 & 24.39 \\
\textbf{FPF-ALL}+ER     & 70.64$\pm$4.00 & 8.79 & 63.81$\pm$2.33 & 34.92 \\
\textbf{FPF}+AGEM       & 73.32$\pm$3.73 & 7.07 & 62.40$\pm$1.89 & 18.47 \\
\textbf{FPF-ALL}+AGEM   & 74.80$\pm$3.12 & 8.79 & 62.65$\pm$1.65 & 34.92 \\
\textbf{FPF}+iCaRL      & 73.56$\pm$3.00 & 4.27 & -              & -\\
\textbf{FPF-ALL}+iCaRL  & 72.77$\pm$4.12 & 8.79 & -              & - \\
\textbf{FPF}+FDR        & 73.64$\pm$3.85 & 2.94 & 65.39$\pm$1.83 & 11.70 \\
\textbf{FPF-ALL}+FDR    & 74.24$\pm$1.48 & 8.79 & 64.88$\pm$2.28 & 34.92 \\
\textbf{FPF}+DER        & 74.13$\pm$3.12 & 2.96 & 65.69$\pm$1.66 & 18.47 \\
\textbf{FPF-ALL}+DER    & 74.54$\pm$3.19 & 8.79 & 66.22$\pm$0.87 & 34.92 \\
\textbf{FPF}+DER++      & 77.37$\pm$1.32 & 4.68 & 66.89$\pm$1.32 & 24.39 \\
\textbf{FPF-ALL}+DER++  & 77.16$\pm$1.45 & 8.79 & 65.19$\pm$1.33 & 34.92 \\
\bottomrule
\end{tabular}
}
\label{tab:FPF_vs_FPFALL}
\end{small}
\end{center}
\end{table*}

\section{Performance of various methods during the training of CL}

In Tab.~\ref{tab:avg_acc_pathmnist} and Tab.~\ref{tab:avg_acc_pacs}, the average test accuracy of previous tasks at the end of each task during the training of CL on Seq-PathMNIST and Seq-PACS is reported. The results show that during training, $k$-FPF can always achieve the best performance among various CL methods. Whenever the training stops, $k$-FPF can always achieve a model performing well on previous tasks.

\begin{table}[ht]
\caption{\footnotesize
The average accuracy of previous tasks at the end of each task during the training of CL on Seq-PathMNIST.
}
\begin{center}
\begin{small}
\resizebox{0.6\textwidth}{!}{
\begin{tabular}{lcccc}
\toprule
\textbf{ Methods}  & Task 1 & Task 2 & Task 3 & Task 4  \\
\midrule
$k$-FPF-CE & 99.95$\pm$0.04 & 95.41$\pm$1.98 & 81.92$\pm$2.26 & 76.72$\pm$1.94  \\
ER         & 98.62$\pm$1.59 & 83.06$\pm$3.12 & 74.60$\pm$3.18 & 57.54$\pm$3.05 \\
AGEM       & 99.71$\pm$0.19 & 46.58$\pm$3.13 & 36.12$\pm$3.17 & 27.33$\pm$3.93 \\
iCaRL      & 99.98$\pm$0.02 & 86.86$\pm$5.47 & 66.62$\pm$5.64 & 57.67$\pm$1.13 \\
FDR        & 99.97$\pm$0.06 & 48.06$\pm$0.82 & 55.75$\pm$6.55 & 40.08$\pm$4.13 \\
DER        & 99.98$\pm$0.02 & 91.92$\pm$3.42 & 76.50$\pm$5.77 & 66.71$\pm$3.40 \\
DER++      & 99.95$\pm$0.06 & 94.06$\pm$6.14 & 80.35$\pm$3.32 & 71.09$\pm$2.60 \\
\bottomrule
\end{tabular}
}
\label{tab:avg_acc_pathmnist}
\end{small}
\end{center}
\end{table}

\begin{table}[ht]
\caption{\footnotesize
The average accuracy of previous tasks at the end of each task during the training of CL on Seq-PACS.
}
\begin{center}
\begin{small}
\resizebox{0.6\textwidth}{!}{
\begin{tabular}{lcccc}
\toprule
\textbf{ Methods}  & Task 1 & Task 2 & Task 3 & Task 4  \\
\midrule
$k$-FPF-CE &  70.94$\pm$2.02 & 73.75$\pm$2.68 & 62.37$\pm$0.49 & 65.90$\pm$0.72 \\
ER         &  56.64$\pm$9.04 & 54.34$\pm$9.44 & 46.79$\pm$8.48 & 52.72$\pm$4.01 \\
AGEM       &  47.34$\pm$7.35 & 38.02$\pm$5.82 & 32.70$\pm$7.13 & 35.29$\pm$4.94 \\
FDR        &  58.59$\pm$4.36 & 54.00$\pm$4.01 & 46.38$\pm$4.80 & 48.50$\pm$4.67 \\
DER        &  48.49$\pm$9.40 & 45.28$\pm$8.88 & 34.48$\pm$7.81 & 47.63$\pm$3.85 \\
DER++      &  55.33$\pm$7.45 & 64.43$\pm$6.50 & 50.19$\pm$7.30 & 57.77$\pm$2.54 \\
\bottomrule
\end{tabular}
}
\label{tab:avg_acc_pacs}
\end{small}
\end{center}
\end{table}



         

\section{Detailed dynamics of BN weights and bias in different groups}
\label{sec:bn_dynamics}
\begin{figure*}[htbp]
     \centering
         \centering
         \includegraphics[width=0.5\columnwidth]{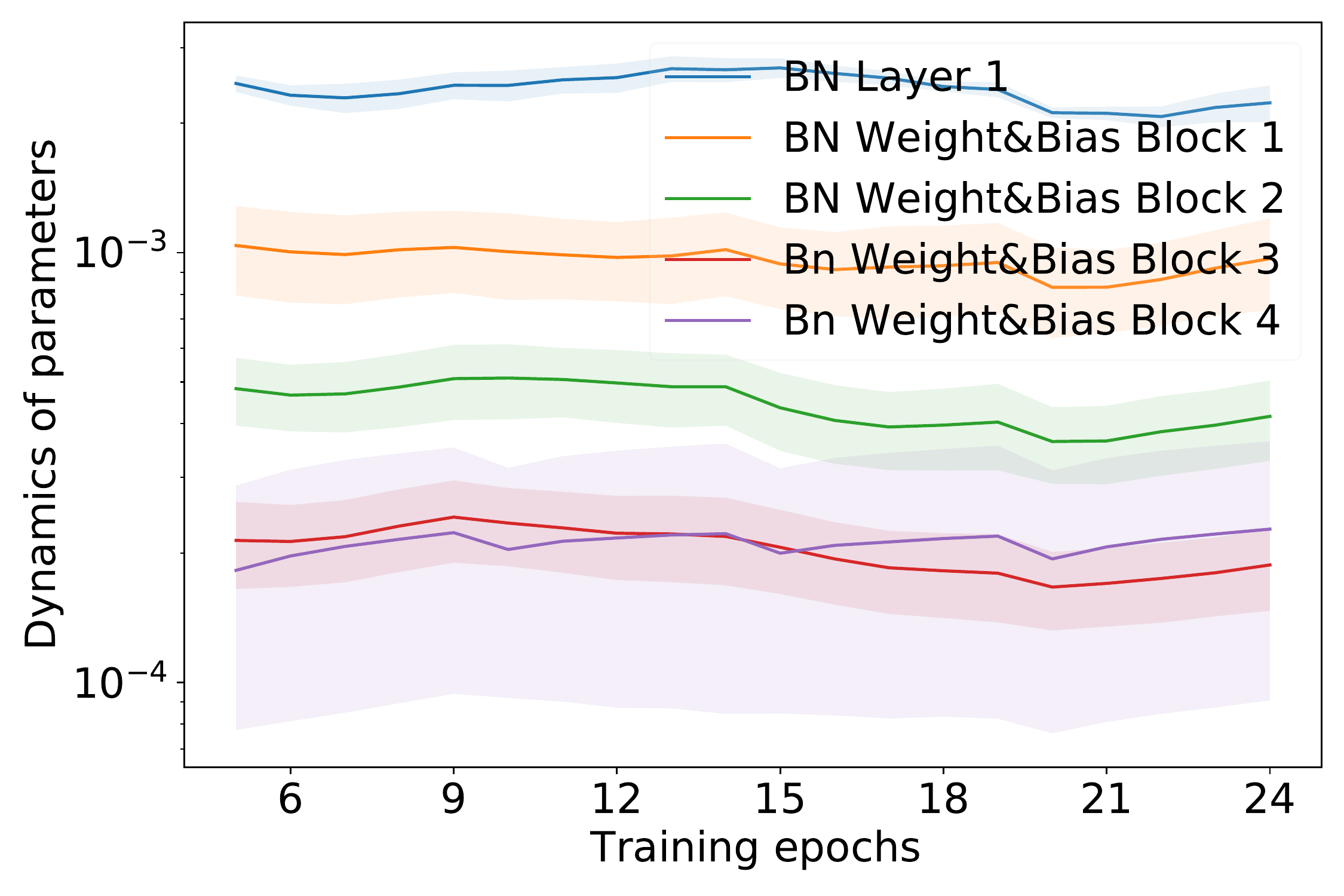}
\caption{\footnotesize The training dynamics of different groups of BN weights and biases in ResNet-18.
}
\label{fig:bn_dynamics}
\end{figure*}

In Fig.~\ref{fig:bn_dynamics}, the training dynamics of BN weights and biases in different groups are reported. This provides a fine-grained explanation of the phenomenon in Fig.~\ref{fig:more_dynamics} (d): the bottom BN layer is much more sensitive and task-specific than other BN layers. Consistent with convolutional layers, the deep BN layers are less sensitive to task drift than the shallower ones. 

In a neural network, lower layers are closer to the input. Since the distribution of the inputs changes, the parameters of lower convolutional layers change sensitively to adapt to the distribution shift. The weights and biases of BN, which are the scale and shift of the featuremap, will change along with the convolutional parameters to adjust the distribution of the output featuremap.
In the deeper layers, the functionality of each filter is relatively stable, so the distribution of the featuremap need not change drastically.

\section{Results of other neural networks}
In Tab.~\ref{tab:results_for_mlpvgg}, the results of various CL benchmarks and FPF on MLP and VGG-11 are reported. Similar to the results in Tab.\ref{tab:main_results}, by finetuning the most sensitive parameters in MLP and VGG-11, FPF can further improve the performance of all SOTA CL methods and achieve the best performance. $k$-FPF-CE also achieves comparable performance as FPF + SOTA methods. Our methods can be generalized to various neural networks.

\begin{table}[htbp]
\caption{
Classification results for CL benchmarks and FPF on MLP and VGG-11.
\textbf{Bold} and \underline{underline} indicate the best and second-best algorithms in each setting.}
\begin{center}
\begin{small}
\begin{sc}
\resizebox{0.6\textwidth}{!}{
\begin{tabular}{clcc}
\toprule
\multirow{2}*{\textbf{Buffer}} & \multirow{2}*{\textbf{Methods}} & \multicolumn{2}{c}{\textbf{Class-IL}}  \\
&  & \multicolumn{1}{c}{Seq-Mnist(MLP)}& \multicolumn{1}{c}{Seq-CIFAR10(VGG-11)}\\
\midrule
& JOINT & 95.58$\pm$0.33 & 69.50$\pm$0.73 \\
& SGD   & 19.64$\pm$0.07 & 18.71$\pm$0.33   \\
& oEWC  & 20.69$\pm$1.34 & 18.46$\pm$0.23  \\
\midrule
\multicolumn{1}{c}{\multirow{13}*{500}}
&  GDUMB                & 90.60$\pm$0.37 & 41.65$\pm$0.78  \\
&  $k$-FPF-CE                & 90.63$\pm$0.57 & 55.45$\pm$1.16  \\
& ER                    & 86.73$\pm$1.03 & 46.27$\pm$1.18 \\
& FPF+ER               & 91.15$\pm$0.16 & 53.48$\pm$1.08 \\
& AGEM                 & 51.03$\pm$4.94  & 19.40$\pm$1.09  \\
& FPF+AGEM            & 89.26$\pm$0.52 & 29.84$\pm$1.37 \\
& iCaRL                 & 58.12$\pm$1.94 & 45.63$\pm$1.94 \\
& FTF+iCaRL            & 80.83$\pm$0.49 & 48.03$\pm$0.65 \\
& FDR                   & 83.79$\pm$4.15  & 45.56$\pm$2.23  \\
& FPF+FDR              & 89.67$\pm$0.37 & \underline{55.59$\pm$1.56}  \\
& DER                   & 91.17$\pm$0.94 & 51.12$\pm$2.47 \\
& FPF+DER              & \textbf{91.25$\pm$0.89} & \textbf{57,46$\pm$1.15} \\
& DER++                 & 91.18$\pm$0.74 & 47.60$\pm$3.23  \\
& FTF+DER++            & \underline{91.22$\pm$0.67} & 54.69$\pm$0.73 \\
\bottomrule
\end{tabular}
}
\label{tab:results_for_mlpvgg}
\end{sc}
\end{small}
\end{center}
\vskip -0.1in
\end{table}

\section{Comparison between $k$-FPF, DER and DER++ with a large number of epochs for each task.}

We compare the accuracy and FLOPs of our methods with the original results in ~\cite{buzzega2020dark} when allowing a large number of epochs on the same data for each task. The results are shown in Tab.~\ref{tab:deroriacc}. R-MNIST is a domain-IL dataset applied in ~\cite{buzzega2020dark}. In both class-IL and domain-IL, k-FPF-CE is comparable to DER++ and k-FPF-KD is better than DER++ on the accuracy but spends much less FLOPs. These results demonstrate that our methods can outperform SOTA methods in various scenarios.

\begin{table}[ht]
\caption{\footnotesize
Comparison of accuracy and FLOPs of $k$-FPF with the original results in ~\cite{buzzega2020dark} which has a large number of training epochs for each task.
}
\begin{center}
\begin{small}
\resizebox{0.8\textwidth}{!}{
\begin{tabular}{lcccc}
\toprule
\textbf{ Methods}  & Seq-CIFAR-10 Accuracy	 & 	Seq-CIFAR-10 FLOPs (B) & R-MNIST Accuracy & R-MNIST FLOPs (B)  \\
\midrule
$k$-FPF-CE &  71.93$\pm$0.58 & 9208.85 & 91.15$\pm$0.29 & 0.64 \\
$k$-FPF-KD &  \textbf{74.32$\pm$0.32} & 9208.85 & \textbf{93.61$\pm$0.45} & 0.64 \\
DER        &  70.51$\pm$1.67 & 16726.26 & 92.24$\pm$1.12 & 1.29 \\
DER++      &  72.70$\pm$1.36 & 25089.39 & 92.77$\pm$1.05 & 1.93 \\
\bottomrule
\end{tabular}
}
\label{tab:deroriacc}
\end{small}
\end{center}
\end{table}

\section{Performance of finetuning different parameters for FPF and k-FPF on domain-IL dataset}
\label{sec:domain_flops}

\begin{figure*}[htbp]
     \centering
         \centering
         \includegraphics[width=0.8\columnwidth]{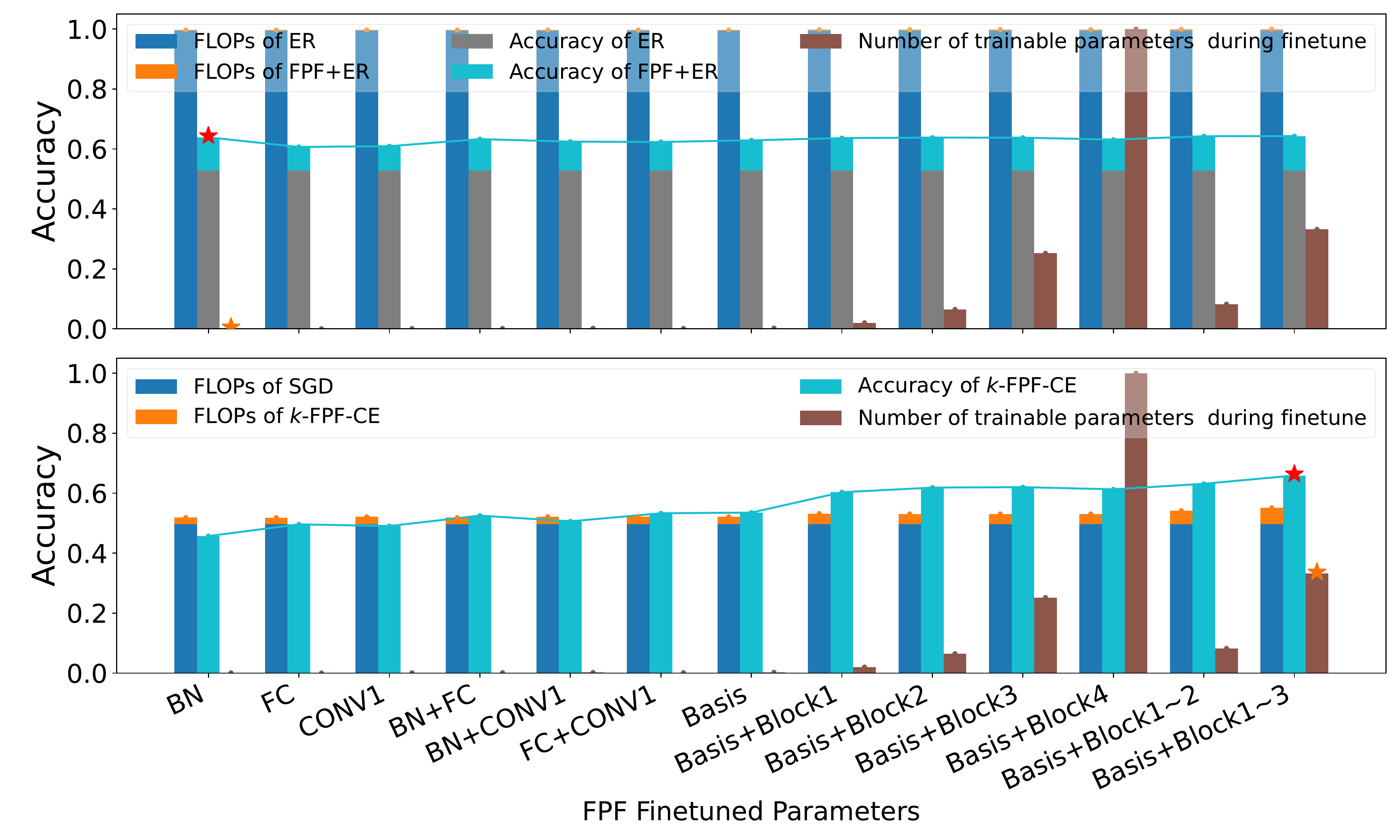}
\caption{\footnotesize Comparison of FLOPs, number of finetuned parameters, and accuracy for FPF(Top) and $k$-FPF(Bottom) finetuning different combinations of parameters. All FLOPs are normalized together to (0,1], as well as the number of finetuning parameters. ``Basis'' in the x-label refers to ``BN+FC+CONV1''.
Red stars highlight the best accuracy and show both FPF and $k$-FPF only require to finetune a small portion of task-specific parameters. $k$-FPF halves FPF's FLOPs.
\textbf{Different from the results of $k$-FPF in class-IL, in Seq-PACS, since the output classes for different tasks are always the same, the last FC layer will not have a large bias on particular classes. Only finetuning BN or CONV1 layers for $k$-FPF can get comparable performance with ER.} Similar to class-IL, since experience replay is not allowed during the training of CL method SGD, a little more parameters are required to be finetuned by $k$-FPF to get comparable performance with FPF (about $24.92\%$ of all parameters).
}
\label{fig:ft_diffparts_flops_pacs}
\end{figure*}

In Figure~\ref{fig:ft_diffparts_flops_pacs}, the performance of finetuning different parameters for FPF and $k$-FPF on domain-IL dataset Seq-PACS are reported. 

\section{Details of datasets}
\label{sec:dataset_detail}

We conduct class-IL experiments on Seq-MNIST, Seq-OrganAMNIST, Seq-PathMNIST, Seq-CIFAR-10, and Seq-TinyImageNet.
Both Seq-OrganAMNIST and Seq-PathMnist consist of 4 disjoint classification tasks. The number of classes per task in Seq-OrganAMNIST and Seq-PathMnist are [3, 3, 3, 2] and [3, 2, 2, 2], respectively.
Seq-MNIST (Seq-CIFAR-10) are generated by splitting the 10 classes in MNIST\cite{lecun1998gradient} (CIFAR-10\cite{krizhevsky2009learning}) into five binary classification tasks. Seq-TinyImageNet partitions the 200 classes of TinyImageNet\cite{le2015tiny} into 10 disjoint classification tasks with 20 classes per task.

For domain-IL experiments, we use PACS dataset~\cite{li2017deeper}, which is widely used for domain generalization. 
Images in PACS come from seven classes and belong to four domains: Paintings, Photos, Cartoons, and Sketches. In Seq-PACS for CL, each task only focuses on one domain and the sequence of tasks is Sketches $\rightarrow$ Cartoons $\rightarrow$ Paintings $\rightarrow$ Photos (increasing the level of realism over time)~\cite{volpi2021continual}.\looseness-1

\section{Comparison with related works \cite{ramasesh2020anatomy}}
Paper ``Anatomy of catastrophic forgetting: Hidden representations and task sementics'' shows that freezing bottom layers had little impact on the performance of the second task.
(i) Their setting is different: our study and most CL methods focus on the performance of ALL tasks. And it is unfair in terms of parameter amount to compare the freezing effects of multiple layers/blocks (e.g., block 1-3) vs. one layer/block.
(ii) Their result is partially consistent with ours since their unfrozen part covers the last layer and many BN parameters, which are the most sensitive/critical part to finetune in our paper. 
(iii) The rest difference is due to our finer-grained study on parameters and on $>2$ tasks, but this paper only studies two tasks and focuses on the second.
Table \ref{table:freezeconv} shows the class-IL accuracy at the end of each task if freezing different single ResNet blocks (bottom to top: block-1 to block-4). At the end of task 2, our observation is the same as this paper and freezing bottom blocks showed little reduction of accuracy. However, at the end of tasks 3-5, their performance drops, and freezing block-1 drops the most. 

\begin{table}[ht]
\setlength{\tabcolsep}{2.pt}
\begin{center}
	\caption{\footnotesize class-IL accuracy of ER at the end of each task on Seq-CIFAR-10}
\begin{tabular}{lccccc}
\toprule
 & Task-1 & Task-2 & Task-3 & Task-4 & Task-5  \\
\midrule
No Freeze       & $ 97.52 \pm 0.23 $ & $ 80.53 \pm 0.80 $ & $ 63.96 \pm 0.51 $ & $ 58.05 \pm 1.91 $ & $ 57.03 \pm 2.29 $ \\
Freeze conv-1   & $ 97.52 \pm 0.23 $ & $ 79.62 \pm 2.75 $ & $ 63.28 \pm 2.13 $ & $ 56.11 \pm 0.61 $ & $ 55.58 \pm 1.31 $ \\
Freeze block-1  & $ 97.52 \pm 0.23 $ & $ 78.88 \pm 3.01 $ & $ 60.07 \pm 0.61 $ & $ 55.49 \pm 0.22 $ & $ 52.75 \pm 1.90 $ \\
Freeze block-2  & $ 97.52 \pm 0.23 $ & $ 78.93 \pm 3.34 $ & $ 63.78 \pm 2.32 $ & $ 56.23 \pm 0.82 $ & $ 56.55 \pm 3.17 $\\
Freeze block-3  & $ 97.52 \pm 0.23 $ & $ 80.37 \pm 2.35 $ & $ 64.31 \pm 2.23 $ & $ 57.21 \pm 0.40 $ & $ 56.52 \pm 0.76 $\\
Freeze block-4  & $ 97.52 \pm 0.23 $ & $ 80.68 \pm 1.53 $ & $ 64.89 \pm 1.00 $ & $ 53.78 \pm 3.37 $ & $ 54.01 \pm 2.07 $\\
\bottomrule
\end{tabular}
	\makeatletter\def\@captype{table}\makeatother
	\label{table:freezeconv}
\end{center}
\end{table}

\clearpage
\section{Distribution of filters' dynamics in different layers of neural networks}
\label{sec:filter_dynamics}
\begin{figure*}[ht]

     \centering
         \centering
         \includegraphics[width=0.85\columnwidth]{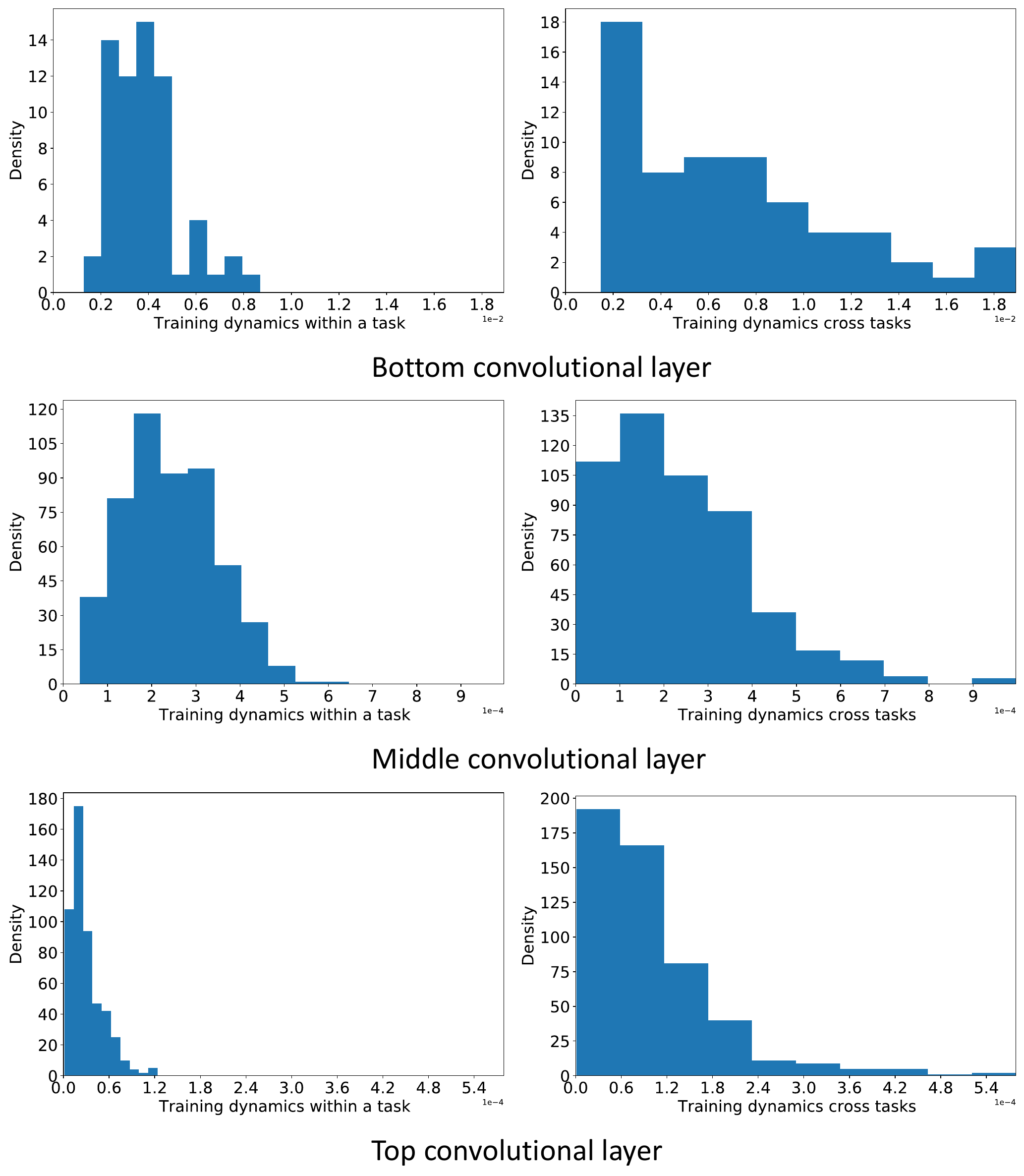}
\caption{\footnotesize Comparison between the distribution of filters' training dynamics within a task and that cross tasks in different convolutional layers of VGG-11. When tasks shift, for all layers, only a small part of the filters in each layer experience more changes.
}
\label{fig:filter_dynamics}
\end{figure*}

\clearpage

\section{A more clear version of Fig.~\ref{fig:methods_flops} and Fig.\ref{fig:ft_diffparts_flops}}
\label{sec:clear_img}
\begin{figure*}[htbp]
     \centering
      \subfigure[Seq-PathMNIST]{
         \centering
         \includegraphics[width=0.85\textwidth]{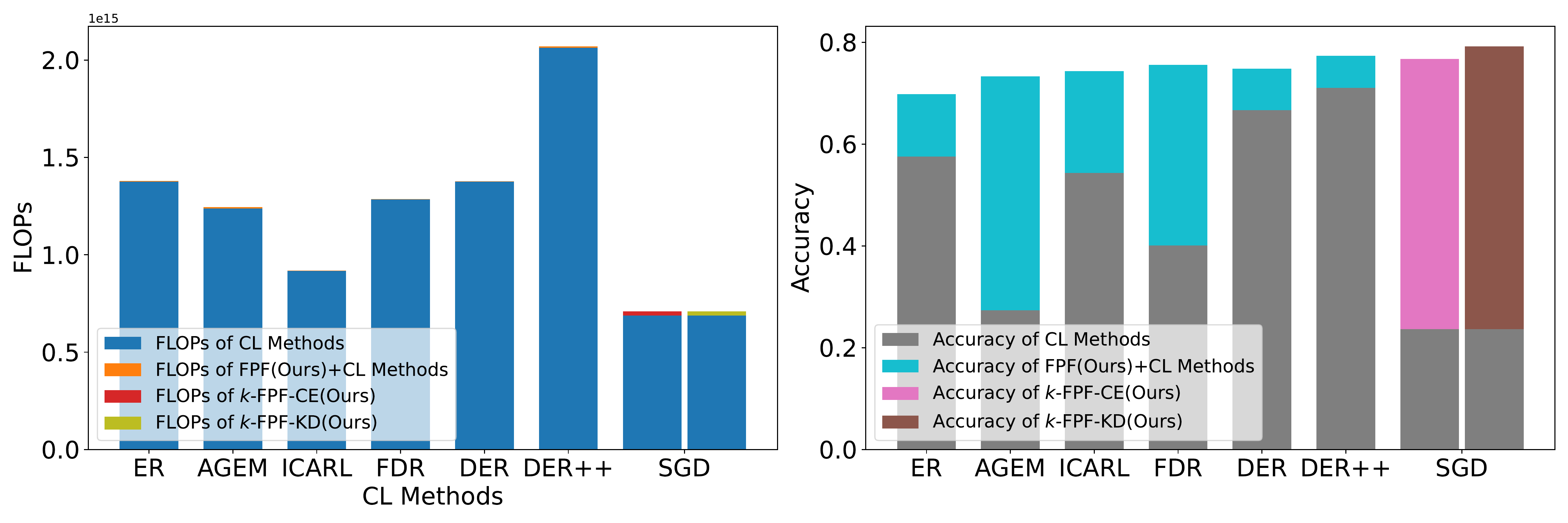}

     }

     \subfigure[Seq-Tiny-ImageNet]{
         \centering
         \includegraphics[width=0.85\textwidth]{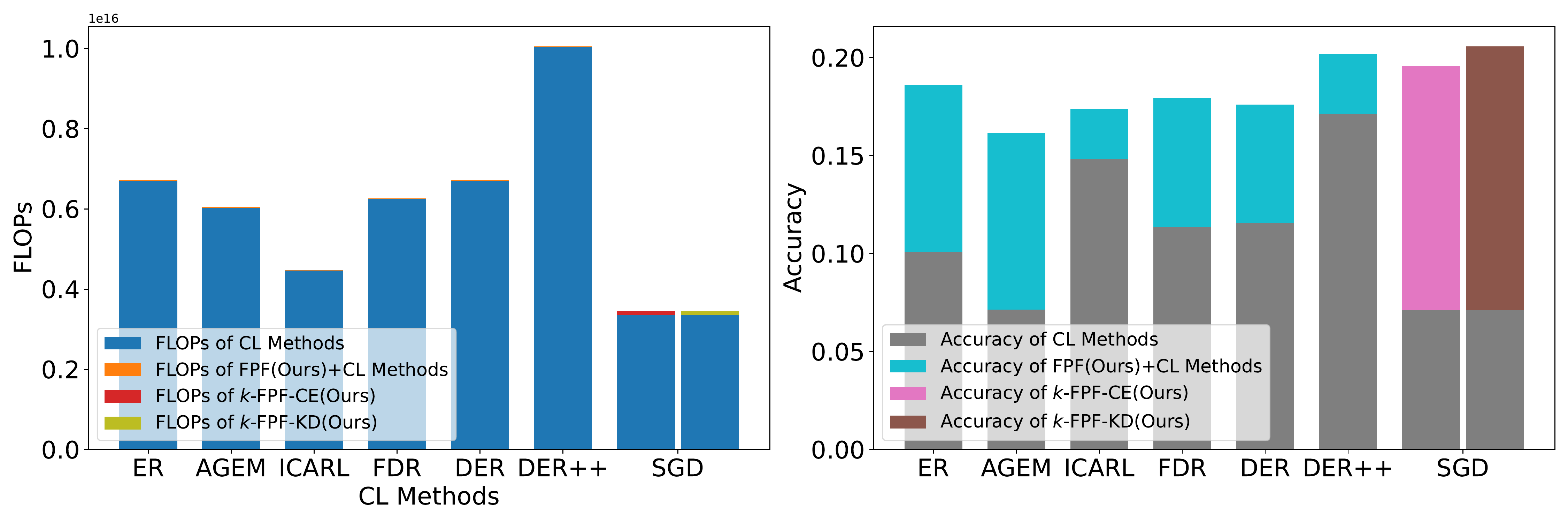}
     }

     \subfigure[Seq-PACS]{
         \centering
         \includegraphics[width=0.85\textwidth]{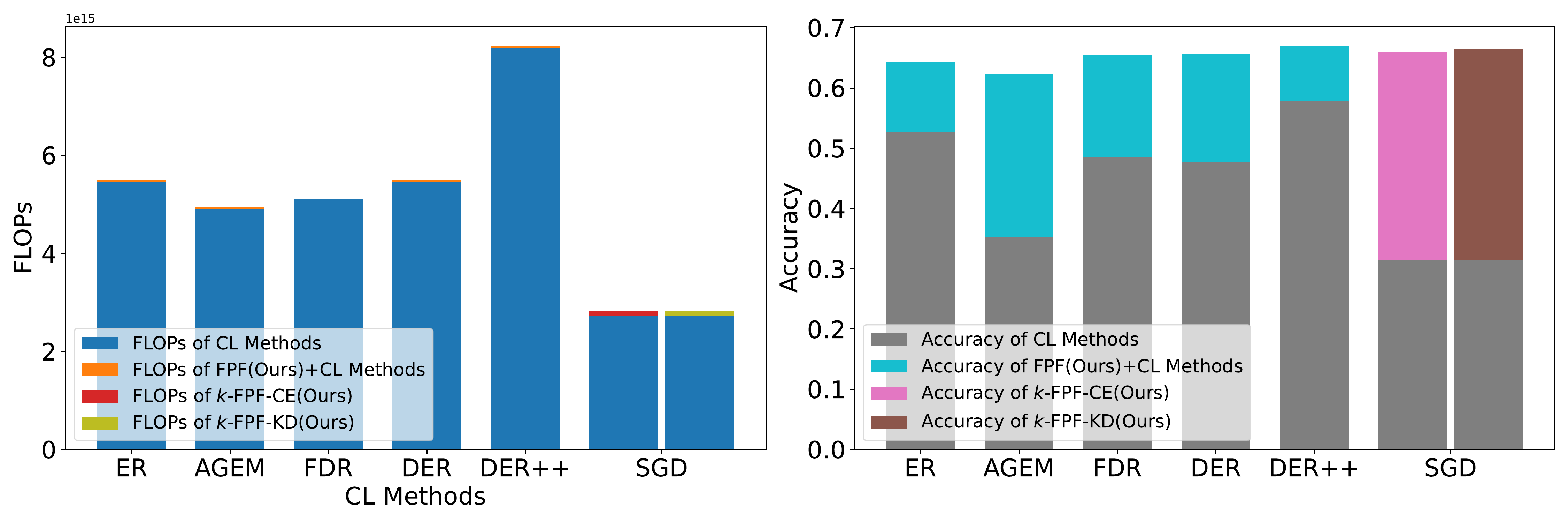}
     }
\caption{\footnotesize Comparison of FLOPs and accuracy between FPF, $k$-FPF and SOTA CL methods. \textbf{FPF improves all CL methods by a large margin without notably extra computation. $k$-FPF consumes much less computation but achieves comparable performance as FPF.}
}
\label{fig:methods_flops_seperate}
\end{figure*}

\begin{figure*}[htbp]

         \centering
         \includegraphics[width=\columnwidth]{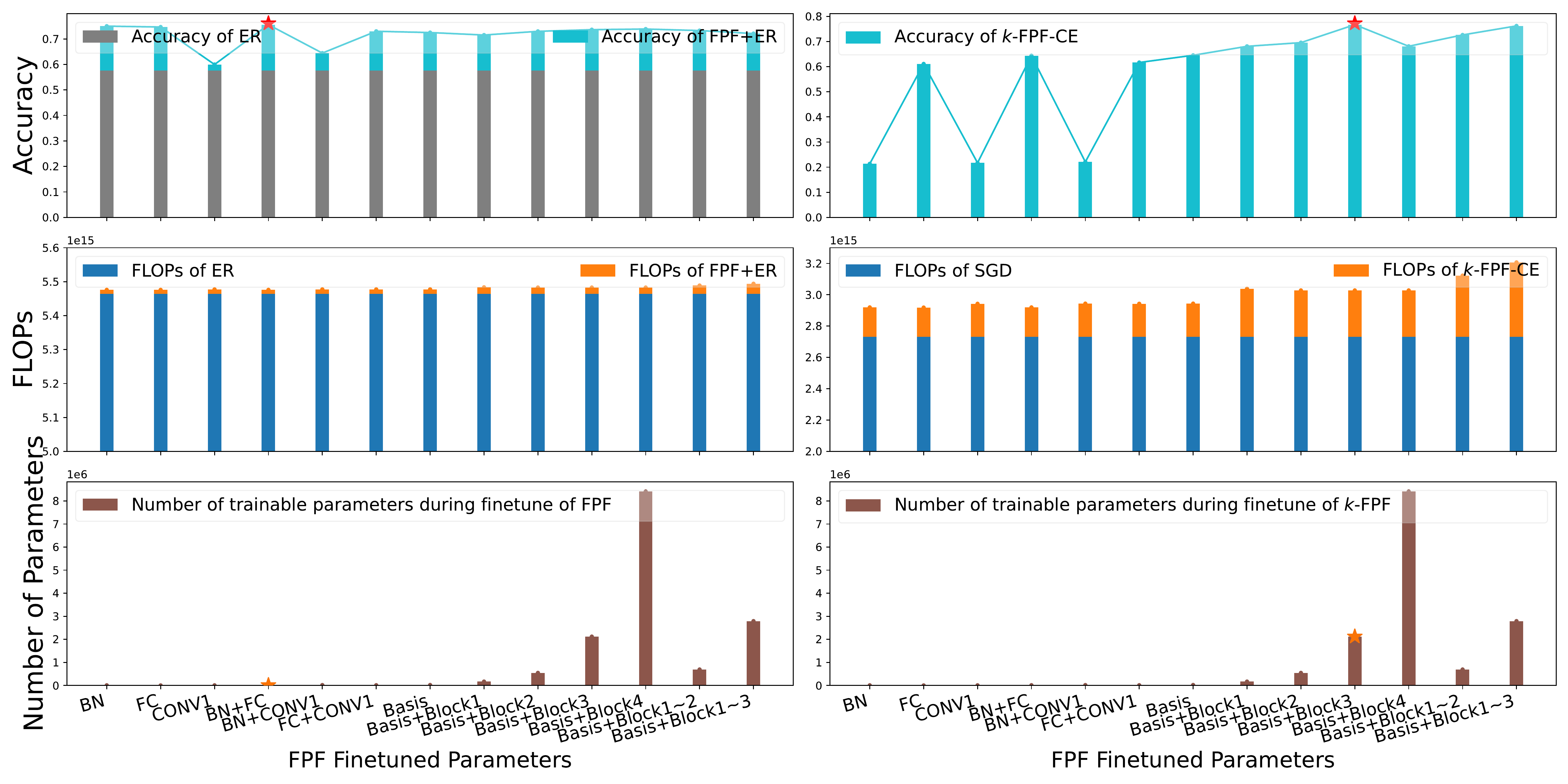}
\caption{\footnotesize Comparison of FLOPs, number of finetuned parameters, and accuracy for FPF(Top) and $k$-FPF(Bottom) finetuning different combinations of parameters. All FLOPs are normalized together to (0,1], as well as the number of finetuning parameters. ``Basis'' in the x-label refers to ``BN+FC+CONV1''.
Red stars highlight the best accuracy and show \textbf{both FPF and $k$-FPF only require to finetune a small portion of task-specific parameters. $k$-FPF halves FPF's FLOPs.} 
}
\label{fig:ft_diffparts_flops_separate}
\end{figure*}

In Fig.\ref{fig:methods_flops_seperate} and Fig.\ref{fig:ft_diffparts_flops_separate}, to make Fig.\ref{fig:methods_flops} and Fig.\ref{fig:ft_diffparts_flops} more concise and easy to understand, we draw the barplots of different parts separately.


\clearpage

\section{Hyper-parameter Search Space}
\label{sec:parameter_search}
In the following, we provide a list of all the hyper-parameter combinations that were considered for FPF and $k$-FPF.

\begin{table*}[h]
\caption{\footnotesize
The hyper-parameter search space for FPF on different datasets.
For all experiments of FPF, we use the same number of batch size 32 and finetuning steps 300. The hyper-parameter spaces of finetuning different parameters in the models generated by different CL methods are always the same for a given dataset. ft-lr refers to the learning rate during finetuning of FPF.
}
\begin{center}
\begin{small}
\resizebox{0.75\textwidth}{!}{
\begin{tabular}{ccl}
\toprule
Dataset & Hyper-parameter & Values \\
\midrule
Seq-OrganAMNIST & lr & [1, 0.3, 0.1, 0.03, 0.01] \\
Seq-PathMNIST & lr & [1, 0.75, 0.3, 0.05, 0.03] \\
Seq-CIFAR-10 & lr & [1, 0.3, 0.1, 0.03, 0.01] \\
Seq-Tiny-ImageNet & lr & [1, 0.5, 0.3, 0.075, 0.05] \\
Seq-PACS & lr & [1, 0.5, 0.3, 0.05, 0.03, 0.005, 0.003] \\
\bottomrule
\end{tabular}
}
\label{tab:FPF_grid_search}
\end{small}
\end{center}
\end{table*}

\begin{table*}[h]
\caption{\footnotesize
The hyper-parameter search space for $k$-FPF-SGD on different datasets.
For all experiments of $k$-FPF-SGD, we use the same number of batch size 32 and finetuning steps 100. The hyper-parameter spaces of finetuning different parameters are always the same for a given dataset. lr refers to the learning rate during training of CL method SGD. ft-lr refers to the learning rate during finetuning.
}
\begin{center}
\begin{small}
\resizebox{0.75\textwidth}{!}{
\begin{tabular}{ccl}
\toprule
Dataset & Hyper-parameter & Values \\
\midrule
Seq-OrganAMNIST & lr & [0.2, 0.15, 0.1, 0.075] \\
                 & ft-lr & [0.5, 0.2, 0.15, 0.1] \\
\midrule
Seq-PathMNIST & lr & [0.05, 0.03, 0.01] \\
                 & lr & [0.1, 0.075, 0.05, 0.03, 0.01] \\
\midrule
Seq-CIFAR-10 & lr & [0.05, 0.03, 0.01] \\
                 & ft-lr & [0.075, 0.05, 0.03, 0.01] \\
\midrule
Seq-Tiny-ImageNet & lr & [0.075, 0.05, 0.03] \\
                 & ft-lr & [0.1, 0.075, 0.05] \\
\midrule
Seq-PACS & lr & [0.05, 0.03, 0.01] \\
                 & ft-lr & [0.075, 0.05, 0.03, 0.0075] \\
\bottomrule
\end{tabular}
}
\label{tab:FPFSGD_grid_search}
\end{small}
\end{center}
\end{table*}

\begin{table*}[h]
\caption{\footnotesize
The hyper-parameter search space for $k$-FPF-KD on different datasets.
For all experiments of $k$-FPF-KD, we use the same number of batch size 32 and finetuning steps 100. The hyper-parameter spaces of finetuning different parameters are always the same for a given dataset. lr refers to the learning rate during training of CL method SGD. ft-lr refers to the learning rate during finetuning. $\lambda$ is the hyper-parameter to balance the two losses.
}
\begin{center}
\begin{small}
\resizebox{0.75\textwidth}{!}{
\begin{tabular}{ccl}
\toprule
Dataset & Hyper-parameter & Values \\
\midrule
Seq-OrganAMNIST & lr & [0.2, 0.15, 0.1, 0.075] \\
                 & ft-lr & [0.5, 0.2, 0.15, 0.1] \\
                 & $\lambda$ & [1, 0.5, 0.2, 0.1] \\
\midrule
Seq-PathMNIST & lr & [0.05, 0.03, 0.01] \\
                 & lr & [0.1, 0.075, 0.05, 0.03, 0.01] \\
                 & $\lambda$ & [1, 0.5, 0.2, 0.1] \\
\midrule
Seq-CIFAR-10 & lr & [0.05, 0.03, 0.01] \\
                 & ft-lr & [0.075, 0.05, 0.03, 0.01] \\
                 & $\lambda$ & [0.5, 0.2, 0.1] \\
\midrule
Seq-Tiny-ImageNet & lr & [0.075, 0.05, 0.03]] \\
                 & ft-lr & [0.1, 0.075, 0.05] \\
                 & $\lambda$ & [1, 0.5, 0.2] \\
\midrule
Seq-PACS & lr & [0.05, 0.03, 0.01] \\
                 & ft-lr & [0.075, 0.05, 0.03, 0.0075] \\
                 & $\lambda$ & [1, 0.5 0.2 0.1] \\
\bottomrule
\end{tabular}
}
\label{tab:FPFKD_grid_search}
\end{small}
\end{center}
\end{table*}

\end{document}